%% file: main.tex
\documentclass[TRO]{RLIpaper}

\newcommand{\e}[1]{ \e{#1}}
\newcommand{\diag}{\text{diag}}

\newcommand{\npoints}{n_{\mathcal{P}}}
\newcommand{\neigen}{n_{M}}
\newcommand{\coveragetime}{t}
\newcommand{\diffusiontime}{\tau}
\newcommand{\nneighbours}{n_{N}}

\usepackage{geometric_algebra}
\usepackage{subfiles}
\usepackage{url}

\author{Cem Bilaloglu$^*$, Tobias Löw$^*$, and Sylvain Calinon%
\thanks{This work was supported by the State Secretariat for Education, Research and Innovation in Switzerland for participation in the European Commission's Horizon Europe Program through the INTELLIMAN project (\url{https://intelliman-project.eu/}, HORIZON-CL4-Digital-Emerging Grant 101070136) and the SESTOSENSO project (\url{http://sestosenso.eu/}, HORIZON-CL4-Digital-Emerging Grant 101070310).} 
\thanks{$^*$Equal contribution. 
The authors are with the Idiap Research Institute, Martigny, Switzerland and with the Ecole Polytechnique Fédérale de Lausanne (EPFL), Switzerland.
        {\tt\footnotesize cem.bilaloglu@idiap.ch; tobias.loew@idiap.ch; sylvain.calinon@idiap.ch}}%
}

\title{Tactile Ergodic Coverage on Curved Surfaces}

\begin{document}

\maketitle

\subfile{sections/0Abstract}

\subfile{sections/1Introduction}

\subfile{sections/2RelatedWork}

\subfile{sections/background}

\subfile{sections/3Method}

\subfile{sections/4Experiments}

\subfile{sections/5Conclusion}

\printbibliography

\begin{IEEEbiography}[{\includegraphics[width=1in,height=1.25in,clip,keepaspectratio]{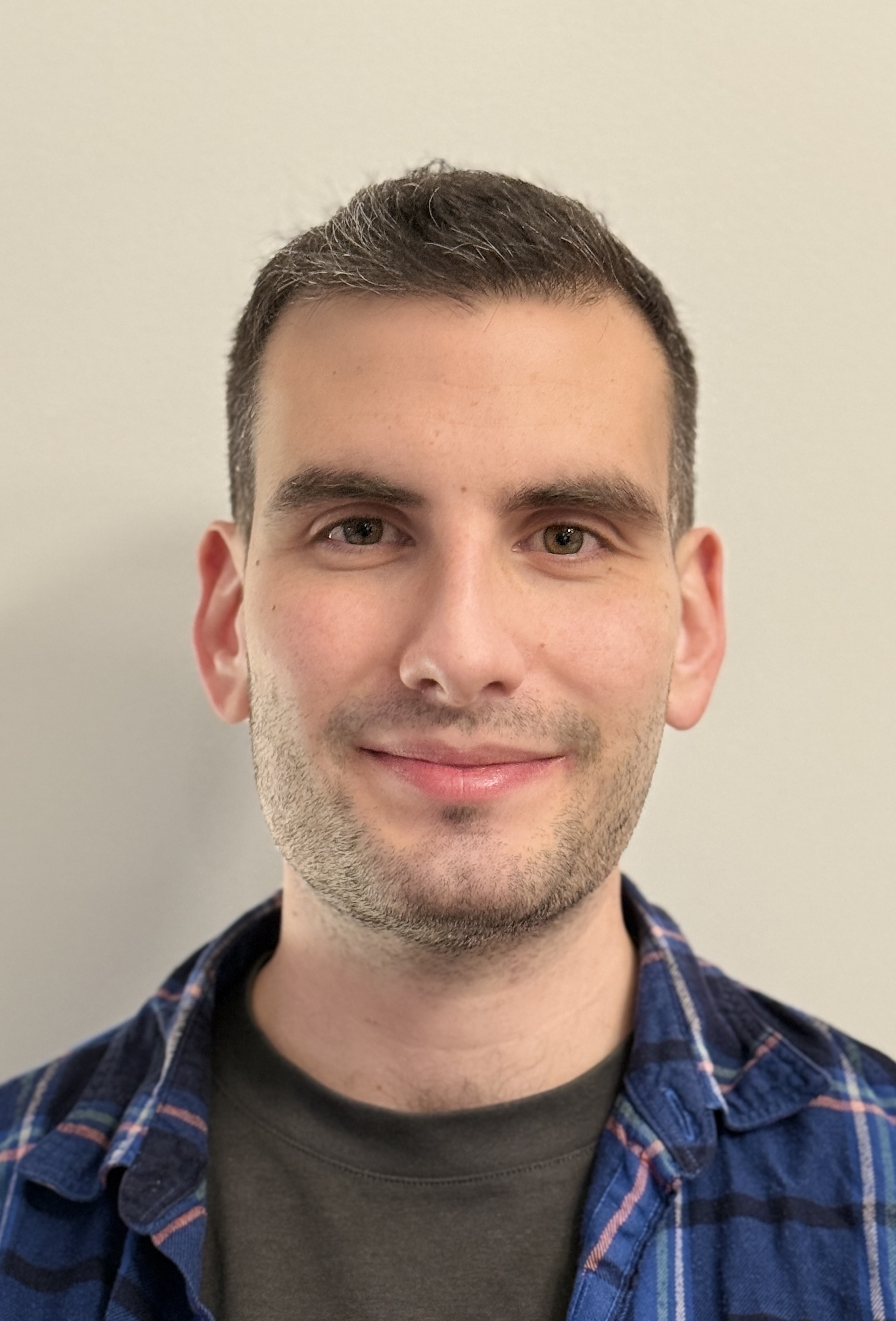}}]{Cem Bilaloglu} is a Ph.D. student in Electrical Engineering at École Polytechnique Fédérale de Lausanne (EPFL) and is working as a research assistant in the Robot Learning and Interaction Group at the Idiap Research Institute. He received his BSc. and MSc. in Mechanical Engineering from METU, Turkey, in 2018 and 2021, respectively. His research focuses on leveraging geometric structures to address challenges in robot learning and control. \newline Website: \url{https://sites.google.com/view/cembilaloglu}.
\end{IEEEbiography}

\begin{IEEEbiography}[{\includegraphics[width=1in,height=1.25in,clip,keepaspectratio]{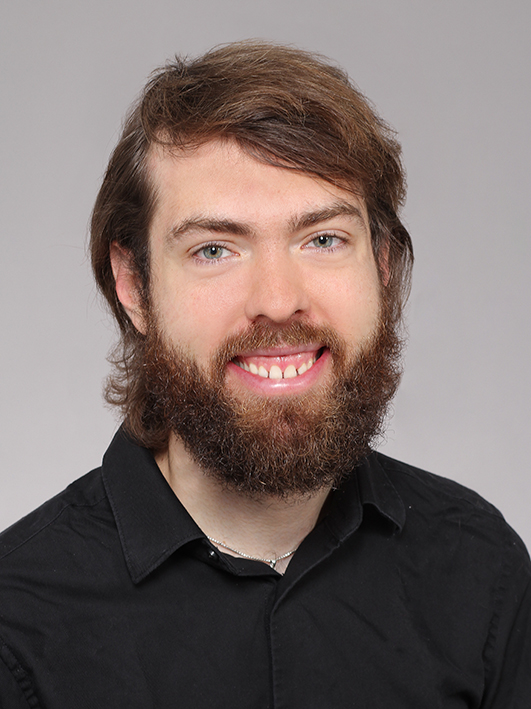}}]{Tobias L\"ow} is a Ph.D. student in Electrical Engineering at École Polytechnique Fédérale de Lausanne (EPFL) and is working as a research assistant in the Robot Learning and Interaction Group at the Idiap Research Institute. He received his BSc. and MSc. in Mechanical Engineering from ETH Z\"urich, Switzerland, in 2018 and 2020, respectively. He conducted his master's thesis at the Robotics and Autonomous Systems Group, CSIRO, Brisbane. His research interests lie in using geometric methods for robot control, learning and optimization problems. \newline Website: \url{https://tobiloew.ch}.
\end{IEEEbiography}

\begin{IEEEbiography}[{\includegraphics[width=1in,height=1.25in,clip,keepaspectratio]{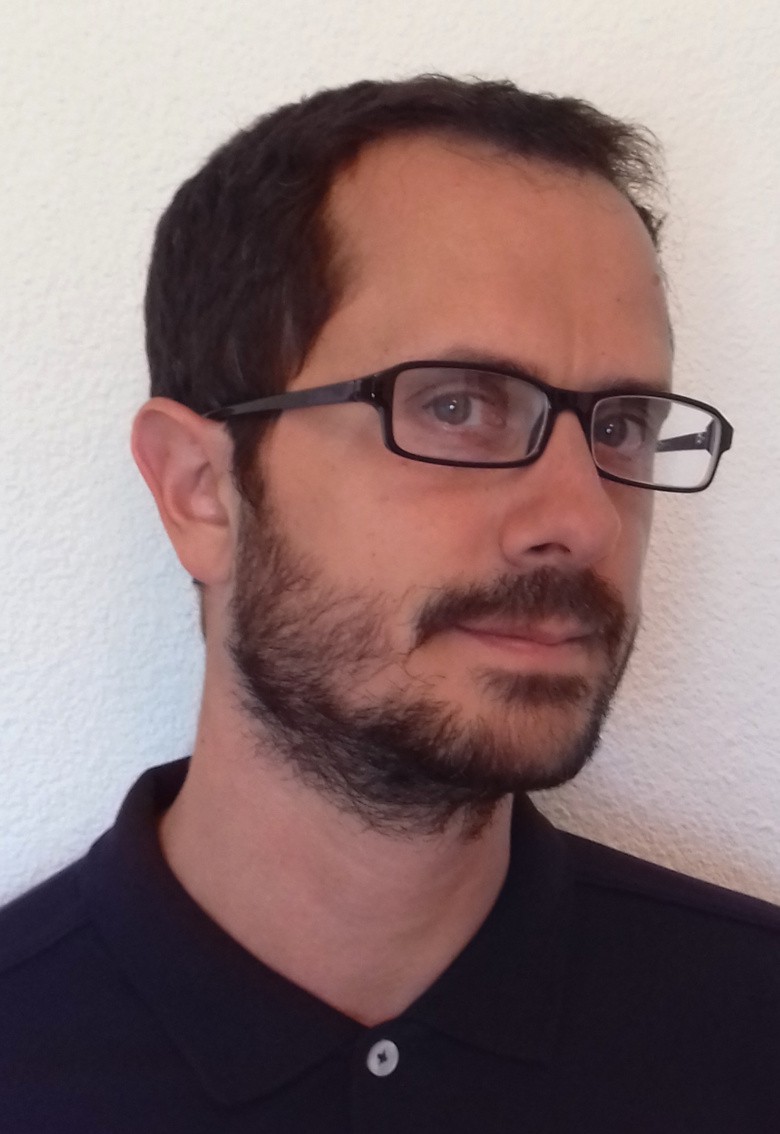}}]{Sylvain Calinon} received the Ph.D. degree in robotics from the École Polytechnique Fedérale de Lausanne (EPFL), Lausanne, Switzerland, in 2007. He is currently a Senior Researcher with the Idiap Research Institute, Martigny, Switzerland, and a Lecturer with the EPFL. From 2009 to 2014, he was a Team Leader with the Italian Institute of Technology, Genoa, Italy. From 2007 to 2009, he was a Postdoc with EPFL. His research interests include human–robot collaboration, robot learning, and model-based optimization. Website: \url{https://calinon.ch}.
\end{IEEEbiography}

\end{document}

%% file: sections/0Abstract.tex
    \begin{abstract}

        
        In this article, we present a feedback control method for tactile coverage tasks such as cleaning or surface inspection. Although these tasks are challenging to plan due to the complexity of continuous physical interactions, the coverage target and progress can be effectively measured using a camera and encoded in a point cloud. We propose an ergodic coverage method that operates directly on point clouds, guiding the robot to spend more time on regions requiring more coverage. For robot control and contact behavior, we use geometric algebra to formulate a task-space impedance controller that tracks a line while simultaneously exerting a desired force along that line. We evaluate the performance of our method in kinematic simulations and demonstrate its applicability in real-world experiments on kitchenware. Our source codes, experiment videos, and data are available  at \url{https://sites.google.com/view/tactile-ergodic-control/}.
    \end{abstract}


%% file: sections/1Introduction.tex

\section{INTRODUCTION}
\label{sec:introduction}

    The long-term vision of robotics is to assist humans with daily tasks. The success of robot vacuum cleaners and lawnmowers as consumer products highlights the potential of robotic assistance for common household chores~\cite{cakmakComprehensiveChoreList2013}. These tasks involve covering a region in a repetitive and exhaustive manner. Currently, these robots are limited to relatively large, planar surfaces, and even navigating slopes remains challenging~\cite{veerajagadheswarSSacrrStaircaseSlope2022, sunSwitchableUnmannedAerial2021}. Other daily tasks, such as washing dishes or grocery items, present even greater challenges due to the complex physical interactions with intricate, curved surfaces. Similarly, numerous coverage tasks on curved surfaces arise in industrial and medical applications. In industrial settings, such tasks include surface operations that remove material, such as sanding~\cite{maricCollaborativeHumanRobotFramework2020}, polishing~\cite{karacanPassivityBasedSkillMotion2022,amanhoudDynamicalSystemApproach2019} or deburring~\cite{onsteinDeburringUsingRobot2020}  as well as surface inspection tasks leveraging contact~\cite{pfandlerNondestructiveCorrosionInspection2024}. In medical settings, similar applications range from mechanical palpation~\cite{ayvaliUsingBayesianOptimization2016a,yanFastLocalizationSegmentation2021} and ultrasound imaging~\cite{jiangPreciseRepositioningRobotic2022, fuOptimizationBasedVariableImpedance2024} to massage~\cite{luoRobotAssistedTapping2017,khoramshahiArmhandMotionforceCoordination2020} and bed bathing~\cite{chih-hungkingAssistiveRobotThat2010,madanRABBITRobotAssistedBed2024}. Last but not least, datasets combining the tactile properties of objects with their shape and visual appearance remain scarce and expensive to collect, as they rely on teleoperation~\cite{linLearningVisuotactileSkills2024}. Thus, tactile coverage is critical for automating the collection of tactile datasets that complement visual ones. The problem definitions of this diverse range of settings and applications can be distilled into two key requirements: (i) tactile interactions with a possibly non-planar surface and (ii) a continuous trajectory of contact points covering a region of interest on the surface. Accordingly, this article addresses the overarching problem of tactile coverage on curved surfaces.
    \begin{figure}[!t]
        \centering
        \includegraphics[width=\linewidth]{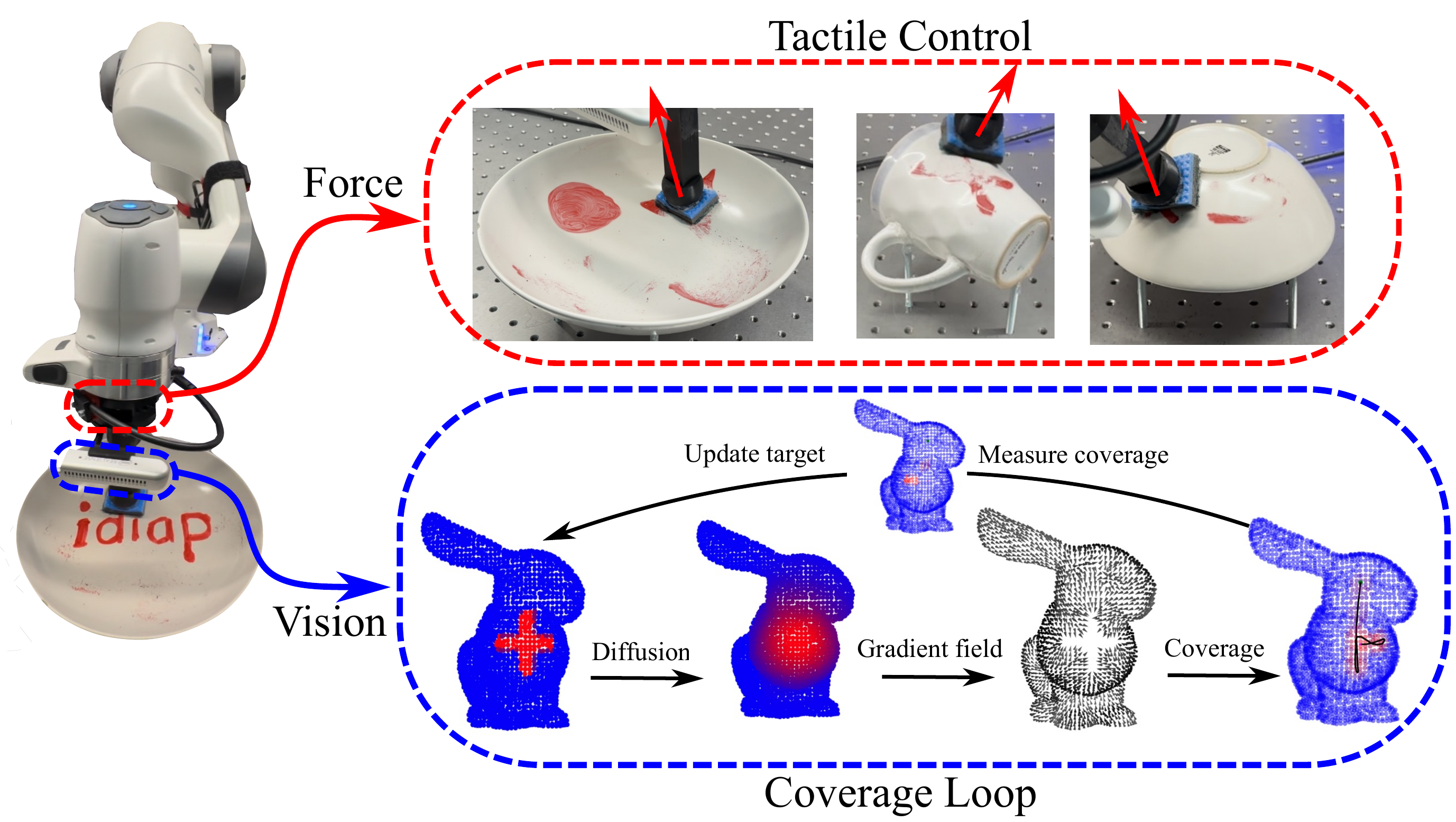}
        \caption{Overview of our feedback control method for tactile coverage. \emph{Left:} We measure the surface and the red target using the camera and encode them in a point cloud. \emph{Bottom-right:} We diffuse the target and use its gradient field to guide the coverage. Then, we close the loop by measuring the actual coverage with the camera and use it as the next target. \emph{Top-right:} We measure the tactile interaction forces using the force sensor and the tool orientation using the joint positions. We solve the geometric task-space impedance control problem using a line target and a force target along the line.}
        \label{fig:overview}
    \end{figure}   

    Tactile tasks involve multiple contact interactions with the environment, making these systems notoriously difficult to control~\cite{aydinogluRealTimeMultiContactModel2022}. While humans solve these tasks effortlessly, they remain extremely challenging for robots. For instance, when cleaning an object, achieving adequate coverage depends on recognizing dirt, understanding the object's material, and assessing their interaction to determine the required contact force for removal. Consequently, the success of coverage depends on unknown or difficult-to-measure parameters, making it challenging to model all interactions. Without an accurate model, motion planning is prone to failure. By analyzing previous research~\cite{kimControlStrategiesCleaning2019} and observing how humans address these challenges, we argue that humans bypass the complexities of planning by solving the simpler closed-loop control problem. Humans leverage visual and tactile feedback for online adaptation. Similarly, robots can measure progress in tactile coverage tasks using vision, turning the task of identifying uncovered regions into an image segmentation problem which has been addressed using various model-based~\cite{kabirAutomatedPlanningRobotic2017,mertensDirtDetectionBrown2005a} or learning-based algorithms~\cite{canedoDeepLearningBasedDirt2021,madanRABBITRobotAssistedBed2024}. However, determining how to control a robot to cover these target regions on curved surfaces remains an open challenge.

    Existing research on coverage has primarily focused on coverage path planning, which involves optimizing a path to ensure that a specified region of interest is covered within a set time frame. Traditionally, the underlying assumption is that visiting each point in the region of interest only once is sufficient for full coverage, an assumption that is reasonable for simple interactions, but not for many tactile tasks. Tactile interactions are often too complex to model deterministically, making it challenging to ensure full coverage after a single visit. Instead, for a cleaning task, a relatively dirty region requires more visits compared to a less dirty region. Similarly, in a surface inspection task, regions requiring higher precision demand more visits to compensate for sensor uncertainty. Furthermore, the robot is expected to keep in contact with the surface while moving, which significantly increases the cost of movement. This cost depends on the geodesic distance on the surface rather than the Euclidean distance. Therefore, naive sampling strategies that fail to account for the cost or constraints of movement and/or surface geometry are unsuitable for tactile coverage tasks. In contrast, \emph{ergodic coverage}~\cite{mathewSpectralMultiscaleCoverage2009} controls the trajectories of dynamical systems by correlating the average time spent in a region to the target spatial distribution. Therefore, ergodic coverage incorporates the motion model as the system dynamics and directly controls the coverage trajectories by using the spatial distribution measured by the vision system.
    
    Considering these challenges, we present a closed-loop tactile ergodic control method that operates on point clouds for tactile coverage tasks. Using point clouds enables us to acquire the target object and spatial distribution at runtime using vision, measure coverage progress, and compensate for unmodeled dynamics in tactile coverage tasks. Our method then constrains the ergodic control problem to arbitrary surfaces to cover a target spatial distribution on the surface. We propagate coverage information by solving the diffusion equation on point clouds, which we compute in real-time by exploiting the surface's intrinsic basis functions called Laplacian eigenfunctions. These eigenfunctions generalize the Fourier series to manifolds (i.e.,\ curved spaces). In order to exert a desired force on the surface while moving, we formulate a geometric task-space impedance controller using geometric algebra. This controller uses surface information to track a line target that is orthogonal to the surface while simultaneously exerting the desired force in the direction of that line. Notably, the geometric formulation ensures that these two objectives do not conflict with each other and can therefore be included in the same control loop without requiring exhaustive parameter tuning. In summary, our proposed closed-loop tactile ergodic control method offers the following contributions:
    \begin{itemize}
        \item formulating the tactile coverage as closed-loop ergodic control problem on curved surfaces;
        \item closing the coverage loop by solving ergodic control problem on point clouds using diffusion; 
        \item achieving real-time frequencies by computing the diffusion using Laplacian eigenfunctions;
        \item contact line and force tracking without conflicting objectives.
    \end{itemize}
    The rest of the article is organized as follows. Section~\ref{sec:related_work} describes related work. Section~\ref{sec:background} provides the mathematical background. Section~\ref{sec:method} presents our method. In Section~\ref{sec:experiments}, we demonstrate the effectiveness of our method in simulated and real-world experiments. Finally, we discuss our results in Section~\ref{sec:discussion}.

%% file: sections/2RelatedWork.tex
\section{RELATED WORK}
\label{sec:related_work}
    The majority of the coverage methods consider the problem from a planning perspective and are generally known as coverage path planning (CPP) algorithms~\cite{zelinskyPlanningPathsComplete1993,latombeRobotMotionPlanning1991,chosetCoverageRoboticsSurvey2001}. Although these methods can handle planar regions with various boundaries~\cite{ollisFirstResultsVisionbased1996, chosetCoveragePathPlanning1998, acarMorseDecompositionsCoverage2002}, their extension to curved surfaces imposes limiting assumptions, such as projectively planar~\cite{hertTerraincoveringAlgorithmAUV1996} or pseudo-extruded surfaces~\cite{atkarHierarchicalSegmentationPiecewise2009}. Additionally, CPP methods assume that the coverage target is uniformly distributed in space. Extending CPP methods to account for spatial correlations in the information leads to informative path planning (IPP)~\cite{zhuOnlineInformativePath2021}. Most IPP and CPP approaches address a variant of the NP-hard traveling salesman problem~\cite{tanComprehensiveReviewCoverage2021}, which limits their scalability as domain complexity increases. Consequently, existing methods are either open-loop~\cite{doGeometryAwareCoveragePath2023} or impose limiting assumptions, such as convexity, for online planning updates~\cite{zhuOnlineInformativePath2021}.

    Closely related to coverage is the problem of exploration, where the environment is initially unknown, and robots gather information using onboard sensors~\cite{acarSensorbasedCoverageExtended2006,shivashankarRealTimePlanningCovering2011}. Tactile exploration is particularly necessary for gathering information on surfaces that can only be acquired through contact~\cite{chittaTactileSensingMobile2011}. A notable example is non-invasive probing (palpation) of tissue stiffness, which aids in disease diagnosis or surgery by providing additional anatomical information. For this purpose, Gaussian processes (GP) have been used for discrete~\cite{ayvaliUsingBayesianOptimization2016a} and continuous~\cite{chalasaniConcurrentNonparametricEstimation2016} probing to map tissue stiffness. While GP-based approaches effectively guide sampling locations, they do not account for the robot's dynamics. This limitation was later addressed by using trajectory optimization to actively search for tissue abnormalities~\cite{salmanTrajectoryOptimizedSensingActive2018}. Unlike other sensing modalities that depend solely on position, tactile interactions also depend on conditions such as relative velocity and contact pressure~\cite{ketchumActiveExplorationRealTime2024a}. To address this, methods have been developed to model forces~\cite{neidhardtOpticalForceEstimation2023} and more complex interactions between robotic tools and surfaces~\cite{vignaliModelingBiomechanicalInteraction2021}. 
    
    The complexity of the problem increases further if we consider scenarios with a robot physically interacting with the environment. For example, in tasks like surface finishing (e.g., polishing, sanding, grinding), the surface itself changes, as material is removed~\cite{ngProgrammingRobotConformance2017}. Similarly, in cleaning tasks, the robot's actions affect the distribution of dirt on the surface~\cite{martinez-hernandezActiveSensorimotorControl2017}. To avoid complex modeling, there are approaches either relying on reinforcement learning~\cite{lewRoboticTableWiping2023} or deep learning~\cite{saitoWiping3DobjectsUsing2020}. In a very similar setting to ours, a manipulator was used to clean the stains on a curved surface by performing multiple passes~\cite{kabirAutomatedPlanningRobotic2017}. However, this work used a sampling-based planner, which required to predefine the maximum number of cleaning passes. In contrast, we relate the target distribution (e.g.,\ stain) directly to feedback control without requiring any task-specific assumptions. 

    In tactile coverage scenarios, visiting a region once can not guarantee full coverage, and predicting how many times the robot should revisit a particular spot is challenging. Consequently, defining a time horizon for trajectory optimization is difficult, as the quality of the result would be significantly affected by this hard-to-make choice. Instead, ergodic control relates how often the robot should revisit a particular spot to the target density at that spot. In this context, \emph{ergodic} describes a dynamical system in which the time averages of functions along its trajectories are equal to their spatial averages~\cite{mathewMetricsErgodicityDesign2011}. The key advantage of ergodic control is its ability to handle arbitrary spatial target distributions without requiring a predefined time horizon. When the spatial target distribution is measurable, the ergodic controller can use this feedback to direct the system to visit regions with higher spatial probabilities more frequently. Recent findings have demonstrated that ergodicity is not merely a heuristic~\cite{berruetaMaximumDiffusionReinforcement2024}; it is the optimal method for collecting independent and identically distributed data while accounting for system dynamics. 

    Ergodic control was introduced in the seminal work by Mathew and Mezić~\cite{mathewSpectralMultiscaleCoverage2009}, which presented the spectral multiscale coverage (SMC) algorithm. SMC is a feedback control law based on the Fourier decomposition of the target distribution and robot trajectories, where \emph{multiscale} aspect prioritizes low-frequency components over high-frequency ones, corresponding to starting with large-scale spatial motions before refining finer details. Since this behavior is achieved through a myopic feedback controller rather than an offline planner, the ergodic controller remains effective even when motion is obstructed~\cite{shettyErgodicExplorationUsing2022}. Recent works have adapted SMC's objective within a trajectory optimization framework to incorporate additional objectives, such as obstacle avoidance~\cite{lerchSafetyCriticalErgodicExploration2023}, time-optimality~\cite{dongTimeOptimalErgodic2023} and energy-awareness~\cite{seewaldEnergyAwareErgodicSearch2024}. Ergodic control has been used for tactile coverage and exploration in applications such as non-parametric shape estimation~\cite{abrahamErgodicExplorationUsing2017} and table cleaning through learning from demonstration~\cite{kalinowskaErgodicImitationLearning2021}. However, all these formulations, which rely on the Fourier decomposition-based ergodic metric, are limited to rectangular domains in Euclidean space.
    
    The first attempt to extend the ergodic control to Riemannian manifolds~\cite{jacobsMultiscaleSurveillanceRiemannian2010} utilized Laplacian eigenfunctions, which generalize the Fourier series to curved spaces. However, this approach was restricted to homogeneous manifolds, such as spheres and tori, where closed-form expressions for the Laplacian eigenfunctions are available. More recently, the \emph{kernel ergodic metric}~\cite{sunFastErgodicSearch2024} was introduced as an alternative to SMC's ergodic metric, enabling extensions to Lie groups and offering improved computational scalability. Nonetheless, arbitrary curved surfaces collected using sensors, such as point clouds, lack both the group structure and the homogeneous manifold properties, presenting additional challenges.

    Another alternative to SMC is the heat equation-driven area coverage (HEDAC) algorithm~\cite{ivicErgodicityBasedCooperativeMultiagent2017}, which uses the diffusion equation, a second-order partial differential equation (PDE), to propagate information about uncovered regions to agents across the domain. Similar to SMC, the original HEDAC implementation was restricted to rectangular domains and lacked collision avoidance. Subsequent extensions have adapted HEDAC to planar meshes with obstacles~\cite{ivicConstrainedMultiagentErgodic2022}, maze exploration~\cite{crnkovicFastAlgorithmCentralized2023}, and CPP on non-planar meshes~\cite{ivicMultiUAVTrajectoryPlanning2023}. However, its application on curved surfaces remains limited to meshes and offline planning due to the heavy pre-processing required.

    In addition to its use in HEDAC, the diffusion equation is widely used in geometry processing tasks, ranging from geodesic computation~\cite{craneDigitalGeometryProcessing2013} to learning on surfaces~\cite{sharpDiffusionNetDiscretizationAgnostic2022}. Its key advantage lies in its ability to account for surface geometry while remaining agnostic to the underlying representation and discretization~\cite{sharpDiffusionNetDiscretizationAgnostic2022}. The diffusion equation is governed by a second-order differential operator called the Laplacian which can be computed for arbitrary surfaces represented as meshes or point clouds using various discretization schemes~\cite{belkinConstructingLaplaceOperator2009,liuPointBasedManifoldHarmonics2012, caoPointCloudSkeletons2010}. In this work, we use a recent approach proposed by Sharp \emph{et al.} which provides a robust and efficient implementation~\cite{sharpLaplacianNonmanifoldTriangle2020}, capable of handling partial and noisy point clouds.

%% file: sections/background.tex
    \section{BACKGROUND}
    \label{sec:background}
        
        
        \subsection{Ergodic Control using Diffusion}
        \label{sub:ergodic_control}
        The ergodic control objective correlates the time that a coverage agent spends in a region to the probability density specified in that region. The HEDAC method~\cite{ivicErgodicityBasedCooperativeMultiagent2017} encodes the coverage objective in the domain $\bm{x} \in \Omega$ at time $\coveragetime$ using a virtual source term
         \begin{equation}{\label{eq:virtual_source}}
            s(\bm{x}, \coveragetime) = \max{\big(p(\bm{x}) - c(\bm{x}, \coveragetime),0\big)}^2,
         \end{equation}
          where $p(\bm{x})$ is the probability distribution corresponding to the coverage target and $c(\bm{x}, \coveragetime)$ is the normalized coverage of the $N$ virtual coverage agents over the domain
         \begin{equation}{\label{eq:normalized_coverage}}
            c(\bm{x}, \coveragetime)=\frac{\tilde{c}(\bm{x}, \coveragetime)}{\int_{\Omega} \tilde{c}(\bm{x}, \coveragetime) d \bm{x}}. 
         \end{equation}
         A single agent's coverage is the convolution of its footprint $\varphi(\bm{r})$ with its trajectory $\bm{x}_i(t^\prime)$. Then, the total coverage becomes the time-averaged sum of these convolutions
        \begin{equation}{\label{eq:coverage}}
            \tilde{c}(\bm{x}, \coveragetime)=\frac{1}{N\coveragetime} \sum_{i=1}^N\int_0^\coveragetime \varphi\big(\bm{x}-\bm{x}_i(t^\prime)\big) d t^\prime.
        \end{equation}
        HEDAC diffuses the source term across the domain $\Omega$ and computes the potential field $u(\bm{x}, \coveragetime)$ using the stationary ($\dot {u}(\bm{x}, \coveragetime)=0$) diffusion 
         \begin{equation}{\label{eq:HEDAC}}
             \alpha \Delta u(\bm{x}, \coveragetime)- u(\bm{x}, \coveragetime) + s(\bm{x}, \coveragetime) = 0, 
         \end{equation}
        with the diffusion coefficient  $\alpha>0$ and the Laplacian operator $\Delta$. The Laplacian is a second-order differential operator which reduces to the sum of the second partial derivatives in Euclidean spaces 
        \begin{equation}
            \Delta f = \nabla \cdot \nabla f =\sum_{i=1}^{n}\frac{\partial^2f}{\partial \bm{x}_i^2}, \quad \forall \bm{x} \in \mathbb{R}^n.
        \end{equation}
        The stationary diffusion~\eqref{eq:HEDAC} governs the potential field within the interior of the domain $\Omega$, while the behavior on the boundary $\partial \Omega$ is dictated by the zero-Neumann boundary condition
        \begin{equation}
           \mathbf{n} \cdot \nabla u(\boldsymbol{x}, \coveragetime)=0, \quad \forall \bm{x} \in \partial \Omega,
        \end{equation}
        where $\mathbf{n}$ represents the outward unit normal vector to the boundary $\partial \Omega$. To guide the $i$-th coverage agent, HEDAC utilizes the smooth gradient field of the diffused potential $u(\bm{x}, \coveragetime)$ and simulates first-order dynamics \cite{lowDrozBotUsingErgodic2022}
        \begin{equation}
            \dot{\bm{x}}_i = \nabla u(\bm{x}_i, \coveragetime).
        \end{equation}
        
        

                
        
        \subsection{Conformal Geometric Algebra}
        \label{sub:geometric_algebra}
            Here, we introduce conformal geometric algebra (CGA) with a focus on the mathematical background necessary to understand the methods used in this article. We will use the following notation throughout the paper: $x$ to denote scalars, $\bm{x}$ for vectors, $\bm{X}$ for matrices, $X$ for multivectors and $\bm{\mathcal{X}}$ for matrices of multivectors.

            The inherent algebraic product of geometric algebra is called the geometric product 
            \begin{equation}\label{eq:geometric_product}
                \geometricproduct,
            \end{equation}
            which (for vectors) is the sum of an inner $\inner$ and an outer $\outer$ product. The inner product is the metric product and therefore depends on the metric of the underlying vector space over which the geometric algebra is built. The underlying vector space of CGA is $\mathbb{R}_{4,1}$, which means there are four basis vectors squaring to 1 and one to -1. The outer product, on the other hand, is a spanning operation that effectively makes subspaces of the vector space elements of computation. These subspaces are called blades. In the case of CGA, there are 32 basis blades of grades 0 to 5. The term grade refers to the number of basis vectors in a blade that are factorizable under the outer product. Vectors, consequently, are of grade 1 and the outer product of two independent vectors, called bivectors, are of grade 2. A general element of geometric algebra is called a multivector. 

            In practice, CGA actually applies a change of basis by introducing the two null vectors $\gae{0}$ and $\gae{\infty}$, which can be thought of as a point at the origin and at infinity, respectively. Since the Euclidean space is embedded in CGA, we can embed Euclidean points $\bm{x}$ to conformal points $P$ via the conformal embedding
            \begin{equation}\label{eq:conformal_embedding}
                P = \mathcal{C}(\bm{x}) = \gae{0} + \bm{x} + \frac{1}{2}\bm{x}^2\gae{\infty}.
            \end{equation}

            In general, geometric primitives in geometric algebra are defined as nullspaces of either the inner or the outer product, which are dual to each other. The outer product nullspace (OPNS) is defined as 
            \begin{equation}\label{eq:opns}
                \outerproductnullspace.
            \end{equation}
            A similar expression can be found for the inner product nullspace. The conformal points are the basic building blocks to construct other geometric primitives in their OPNS representation. The relevant primitives for this work are lines
            \begin{equation}\label{eq:ga_line_construction}
                L = P_1 \outer P_2 \outer \gae{\infty},
            \end{equation}
            which can be constructed from two points and a point at infinity, planes
            \begin{equation}\label{eq:ga_plane_construction}
                E = P_1 \outer P_2 \outer P_3 \outer \gae{\infty},
            \end{equation}
            which can be constructed from three points and a point at infinity and spheres
            \begin{equation}\label{eq:ga_sphere_construction}
                S = P_1 \outer P_2 \outer  P_3 \outer P_4,
            \end{equation}
            which can be constructed from four points.

            Rigid body transformations in CGA are achieved using motors $M$, which are exponential mappings of dual lines, i.e. bivectors (essentially, the screw axis of the motion). Note that motors can be used to transform any object in the algebra, i.e. they can directly be used to transform the previously introduced points, lines, planes and spheres, by a sandwiching operation 
            \begin{equation}\label{eq:motor_sandwich}
                X' = MX\reverse{M},
            \end{equation}
            where is $\reverse{M}$ is the reverse of a motor. 

            The forward kinematics of serial kinematic chains can be found as the product of motors, i.e.
            \begin{equation}\label{eq:forward_kinematics}
                M(\posjoint) = \prod_{i=1}^N M_i(q_i) = \prod_{i=1}^N \exp(q_iB_i),
            \end{equation}
            where $\bm{q}$ is the current joint configuration and $B_i$ are screw axes of the joints. The geometric Jacobian $\gagjacobian\in \mathbb{B}^{1\times N}\subset \cga^{1\times N}$ is a bivector valued multivector matrix and can be found as 
            \begin{equation}\label{eq:ee_geometric_jacobian}
                \gamatrix{J}_G = \mat{
                    B_1'
                    & \ldots &
                    B_N'
                },
            \end{equation}
            where the bivector elements can be found as
            \begin{equation}\label{eq:jacobian_bivector_element}
                B_i' = \prod_{j=1}^i M_j(q_j) B_i \prod_{j=1}^i \reverse{M}_j(q_j).
            \end{equation}

            Twists $\twist$ and wrenches $\wrench$ are also part of the algebra and hence both can be transformed in the same manner as the geometric primitives using \eqref{eq:motor_sandwich}. Note that, contrary to classic matrix Lie algebra, no dual adjoint operation is needed to transform wrenches. There is, however, still a duality relationship between twists and wrenches, which can be found via multiplication with the conjugate pseudoscalar $I_c = I\gae{0}$~\cite{hestenesNewToolsComputational2010}. Both twists and wrenches are bivectors and the space of wrenches can be found as 
            \begin{equation}\label{eq:wrench_space}
                \wrench \in \text{span}\{\gae{23},\gae{13},\gae{12},\gae{01},\gae{02},\gae{03}\}.
            \end{equation}
            The inner product of twists and wrenches $\twist\inner\wrench=-p$ yields a scalar, where $p$ is the power of the motion. Similarly, the inner product of a screw axis and a wrench $B \inner \wrench = - \tau$ yields a torque $\tau$, which we will use for the task-space impedance control in this article.

%% file: sections/3Method.tex

\section{METHOD}
\label{sec:method}
    We present our closed-loop tactile ergodic coverage method in three parts: (i) surface preprocessing; (ii) tactile coverage; and (iii) robot control. The surface preprocessing computes the quantities that need to be calculated only once when the surface is captured. Tactile coverage generates the motion commands for the virtual coverage agent using the precomputed quantities from the surface preprocessing and the robot controller tracks the generated motion commands with a manipulator using impedance control.

    
    \subsection{Problem Statement}
    \label{sub:problem_statement}
        We formulate a tactile ergodic controller that covers target spatial distributions on arbitrary surfaces. Similarly to HEDAC, we propagate the information encoding the coverage objective by diffusing the source term. However, we utilize the non-stationary ($\dot{u}\neq0$) diffusion equation
        \begin{equation}\label{eq:heat}
           \dot{u} (\bm{x}, \tau)= \Delta_{\mathcal{M}} u(\bm{x},\tau),
        \end{equation}
        as it allows control over the desired smoothness \cite{bilalogluWholeBodyErgodicExploration2023}. Since the diffusion equation depends on time, we introduce an additional time variable $\diffusiontime$. The diffusion time $\diffusiontime$ is independent of the coverage time $\coveragetime$ used by the HEDAC algorithm and unlike HEDAC, we require an initial condition $u({\bm{x},0})$. We set the initial condition using the source term given in Equation \eqref{eq:virtual_source} which encodes the coverage objective at the $t$-th timestep of the coverage, i.e., $u(\bm{x},0)=s({\bm{x},\diffusiontime})$. Additionally, here we use $\Delta_{\mathcal{M}}$, which generalizes the Laplacian for Euclidean spaces $\Delta$ to non-Euclidean manifolds $\mathcal{M}$. This operator $\Delta_{\mathcal{M}}$ is also known as Laplace-Beltrami operator but for conciseness we will use the term Laplacian. 
        
        Our coverage domains are curved surfaces (i.e. 2-manifolds) and we capture the underlying manifold $\mathcal{M}$ as a point cloud $\mathcal{P}$ composed of $\npoints$ points using an RGB-D camera
        \begin{equation}{\label{eq:point_cloud}}
            \mathcal{P} := \left\{ (\bm{x}_i, \bm{c}_i) \ \middle| \
            \begin{aligned}
            &\bm{x}_i \in \mathbb{R}^3, \ \bm{c}_i \in \{0, \ldots, 255 \}^3 \\
            &\text{for} \ i = 1, \ldots, \npoints
            \end{aligned}
            \right\},
        \end{equation}
        where $\bm{x}_i$ is the position of the $i$-th surface point in Euclidean space and $\bm{c}_i$ is the vector of RGB color intensities. We assume there is a processing pipeline (i.e.,\ such as~\cite{kabirAutomatedPlanningRobotic2017,sharpDiffusionNetDiscretizationAgnostic2022,caronEmergingPropertiesSelfSupervised2021}) which maps the point positions and colors to the probability mass $p_i$ of the spatial distribution encoding the coverage objective. Accordingly, our coverage target becomes a discrete spatial distribution $p(\bm{x}_i)=p_i$ on the point cloud $\mathcal{P}$. 
        
        In order to solve \eqref{eq:heat} on irregular and discrete domains, such as point clouds, we discretize the problem in space and time. Hence, we use $u_{i,\diffusiontime}$ to denote the value of the potential field at the $i$-th point at the $\diffusiontime$-th timestep. We omit the subscript $i$ if we refer to all points.


    \subsection{Surface Preprocessing}
    \label{sub:surface_preprocessing}   

        First, we compute the spatial discretization of the Laplacian $\Delta_{\mathcal{M}}$. Note that there are various approaches for discretizing the Laplacian on point clouds~\cite{sharpLaplacianNonmanifoldTriangle2020,belkinConstructingLaplaceOperator2009, liuPointBasedManifoldHarmonics2012, caoPointCloudSkeletons2010}. In this work, we follow the approach presented in \cite{sharpLaplacianNonmanifoldTriangle2020} and show a simplified version of it here, but refer the readers to the original work for more details. Using this method, the discrete Laplacian is represented by the matrix $\bm{L} \in \mathbb{R}^{\npoints \times \npoints}$
        \begin{equation}{\label{eq:discrete_Laplacian}}
            \bm{L}=\bm{M}^{-1}\bm{C}, 
        \end{equation}
        where $\bm{M}$ is the diagonal mass matrix and $\bm{C}$ is a sparse symmetric matrix called the weak Laplacian. The entries of $\bm{M}$ correspond to the Voronoi cell areas in the local tangent plane around each point of $\mathcal{P}$. Similarly, the entries of $\bm{C}$ are determined by the connectivity of the points on the local tangent space and the distance between the connected points. Note that the local tangent space structure also identifies the boundary points. For a given point, the lines between the original point and its neighbors are constructed. If the angle between two consecutive lines is greater than $\pi/2$, the point is a boundary and its boundary condition is set as zero-Neumann.

        Next, we discretize the diffusion equation~\eqref{eq:heat} in time and incorporate the discrete Laplacian $\bm{L}$. Using the backward Euler method, we derive the implicit time-stepping equation, which remains stable for any timestep $\diffusiontime$
        \begin{equation}{\label{eq:implicit_time_stepping_pre}}
            \frac{1}{\diffusiontime} (\bm{u}_{\diffusiontime}-\bm{u}_{0}) =\bm{L} \bm{u}_{\diffusiontime},
        \end{equation} 
        where $\bm{u}_0$ and $\bm{u}_\diffusiontime$ are column vectors containing the potential field values at the vertices of the point cloud at the initial and final times, respectively. Then, combining  \eqref{eq:discrete_Laplacian} and \eqref{eq:implicit_time_stepping_pre} and solving for $\bm{u}_{\diffusiontime}$ we obtain the linear system
        \begin{equation}{\label{eq:implicit_time_stepping}}
            \bm{u}_{\diffusiontime} =(\boldsymbol{M}-\diffusiontime\  \boldsymbol{C})^{-1} \boldsymbol{M}\bm{u}_0.
        \end{equation}
        Note that solving~\eqref{eq:implicit_time_stepping} requires inverting a large sparse matrix, which might be computationally expensive depending on the size of the point cloud and requires the timestep to be set before the inversion. Alternatively, we can solve the problem in the spectral domain by projecting the original problem and reprojecting the solution back to the point cloud. This procedure generalizes using the Fourier transform for solving the diffusion equation on a rectangular domain in $\mathbb{R}^n$ to arbitrary manifolds. Note that the Fourier series are the eigenfunctions of the Laplacian $\Delta$ in $\mathbb{R}^n$. Therefore, we can use the eigenvectors of the discrete Laplacian $\bm{L}$ for solving the diffusion equation on point clouds.
        
        We can write the generalized (i.e.,\ $\bm{M}\neq\bm{I}$) eigenvalue problem for the Laplacian as
        \begin{equation}{\label{eq:eigenproblem}}
            \bm{C} \bm{\phi}_m=\lambda_m \bm{M} \bm{\phi}_m,
        \end{equation}
        where $\{\lambda_m, \bm{\phi}_m\}$ are the eigenvalue/eigenvector pairs. Since $\bm{M}$ is diagonal and $\bm{C}$ is symmetric positive definite, by the spectral theorem, we know that the eigenvalues are real, non-negative, and in ascending order analogous to the frequency. Therefore, we can use the first $\neigen$ eigenvalue/eigenvector pairs as a low-frequency approximation of the whole spectrum. Furthermore, the eigenvectors are orthonormal with respect to the inner product defined by the mass matrix $\bm{M}$. Accordingly, we can stack the first $\neigen$ eigenvectors $\bm{\phi}_m$ as column vectors to construct the matrix $\bm{\Phi} \in \mathbb{R}^{\npoints\times \neigen}$ encoding an orthonormal transformation $\bm{\Phi}^{\trsp} \bm{M} \bm{\Phi}=\bm{I}$. Then, we can transform the coordinates (shown with superscripts) from the point cloud to the spectral domain
        \begin{equation}{\label{eq:spectral_projection}}
            \bm{u}^{\bm{\phi}}=\bm{\Phi}^\trsp \bm{M} \bm{u}^{\bm{x}}.
        \end{equation}
        Note that this step is equivalent to computing the Fourier series coefficients of a target distribution in SMC. Due to the orthonormal transformation, the PDE on the point cloud becomes a system of decoupled ODEs in the spectral domain. It is well known that the solution of a first-order linear ODE $\dot{x}(\diffusiontime) = -cx(\diffusiontime)$ is given by $x(\diffusiontime) = e^{-c\diffusiontime}x(0)$, where $c$ is a constant and $x(0)$ is the initial condition. Therefore, the solution of the system of ODEs in the spectral domain is given in matrix form as
        \begin{equation}{\label{eq:spectral_diffusion}}
            \bm{u}^{\bm{\phi}}_{\diffusiontime}= \left[
                \begin{array}{ccc}
                    e^{-\lambda_1 \diffusiontime} & \ldots & e^{-\lambda_m \diffusiontime}
                \end{array}
                \right]^\trsp \odot
                \bm{u}^{\bm{\phi}}_0,
        \end{equation}
        where $\odot$ denotes the Hadamard product. We observe from~\eqref{eq:spectral_diffusion} that the exponential terms with larger eigenvalues (i.e.,\ higher frequencies) will decay faster. Therefore, approximating the diffusion using the first $\neigen$ components introduces minimal error. Secondly, similar to the mixed norm used in SMC, the low-frequency spatial features are prioritized. Next, we transform the solution back to the point cloud to get the diffused potential field
        \begin{equation}{\label{eq:point_cloud_projection}}
            \bm{u}^{\bm{x}}=\bm{\Phi} \bm{u}^{\bm{\phi}}.
        \end{equation}
        We can combine~\eqref{eq:spectral_projection},~\eqref{eq:spectral_diffusion} and~\eqref{eq:point_cloud_projection} into a unified spectral scheme
        \begin{equation}{\label{eq:spectral_time_stepping}}
            \bm{u}_{\diffusiontime}=\Phi \left[
                {\begin{array}{ccc}
                    e^{-\lambda_1 \diffusiontime}&
                    \ldots & e^{-\lambda_m \diffusiontime}
                \end{array}}
                \right]^\trsp
                \odot \left(\Phi^\trsp\bm{M}\bm{u}_0\right).
        \end{equation}
        We omit the superscripts when working on the point cloud for brevity. Note that $\diffusiontime$ is the only free parameter in the diffusion computation. However, its value should be adapted according to the mean spacing between the adjacent points $h$ on the point cloud. For that purpose, we introduce the hyperparameter $\alpha>0$ and embed it into the timestep calculation
        \begin{equation}
            \diffusiontime = \alpha h^2.
        \end{equation}
        Accordingly, we can control the diffusion behavior independently of the point cloud size. Increasing $\alpha$ results in longer diffusion times and attenuates the high-frequency spatial features (see~\eqref{eq:spectral_diffusion} for details). This corresponds to a more global coverage~\cite{bilalogluWholeBodyErgodicExploration2023}. Conversely, decreasing $\alpha$ results in shorter diffusion times, which leads to preserving the high-frequency spatial features, hence more local coverage behavior.

        Note that the Laplacian is determined completely by the connectivity on the local tangent space and the distance between these connected points. Therefore, it is invariant to distance preserving (i.e.,\ isometric) transformations such as rigid body motion or deformation without stretching. Accordingly, we compute $\bm{C}$, $\bm{M}$ and derived quantities only once in the preprocessing step for a given surface. Recomputation is not necessary if the object stays still, moves rigidly, or the target distribution $p_i$ changes. 
        



        \subsection{Tactile Ergodic Coverage}
        \label{sub:tactile_coverage}
            We model the actual coverage tool/sensor as a compliant virtual coverage agent shaped as a disk with radius $r_a$. Notably, one can represent arbitrary tool/sensor footprints as a combination of disks~\cite{bilalogluWholeBodyErgodicExploration2023}. We position our agent at the end-effector of our manipulator. Thus, for a given kinematic chain and joint configuration $\bm{q}$, we can use the forward kinematics to compute the position of our agent as a conformal point $P_a$ 
            \begin{equation}\label{eq:current_ee_position}
                P_{a} = M(\posjoint) \gae{0} \reverse{M}(\posjoint).
            \end{equation}
            Since the point cloud is discrete and the agent should move continuously on the surface, we project our agent $P_a$ and its footprint to the closest local tangent space on the point cloud.

            \subsubsection{Local Tangent Space and Coverage Computation}
            \label{sub:projection} 
                Given the agent's position $P_{a}$, we first compute the closest tangent space on the point cloud. For that, we query a K-D tree $\mathcal{T(P)}$ for the points $\bm{x}_i \in \mathcal{P}$ that are within the radius $r_a$ of the agent. 
                Then, we compute the conformal embeddings $P_i$ of the neighboring Euclidean points $\bm{x}_i$ using \eqref{eq:conformal_embedding}. We refer to the set composed of points $P_i$ as the local neighborhood. Then, we fit a tangent space to the local neighborhood by minimizing the classical least squares objective
                \begin{equation}\label{eq:surface_fitting_minimization}
                    \min \sum_{i=1}^{\nneighbours} (P_i \inner \dual{X})^2, 
                \end{equation}
                where $\dual{X}$ is the dual representation of either a plane or a sphere and the inner product $\inner$ is a distance measure. In CGA, planes can be seen as limit cases of spheres, i.e.\ planes are spheres with infinite radius. This is also easy to observe by looking at Equations~\eqref{eq:ga_plane_construction} and~\eqref{eq:ga_sphere_construction} which construct these geometric primitives.  Note that fitting a local tangent sphere with the radius determined by the local curvature would always result in smaller or equal residuals than fitting a plane.
                 
                It has been shown in~\cite{hildenbrandFoundationsGeometricAlgebra2013} that the solution to the least squares problem given in~\eqref{eq:surface_fitting_minimization} is the eigenvector corresponding to the smallest eigenvalue of the $5\times5$ matrix 
                \begin{equation}
                    b_{j,k} = \sum_{i=1}^{\nneighbours} w_{i,j}w_{i,k},
                \end{equation}
                where 
                \begin{equation}
                    w_{i,k} = 
                    \begin{cases}
                        p_{i,k} \hspace{3mm} \text{if } k \in \left\{ 1,2,3 \right\} 
                        \\
                        -1 \hspace{3mm} \text{if } k=4
                        \\
                        - \frac{1}{2}\bm{p}_i^2 \hspace{3mm} \text{if } k=5.
                    \end{cases}
                \end{equation}
                Using the five components $v_i$ of this eigenvector we can find the geometric primitive as
                \begin{equation}\label{eq:surface_primitive}
                    X = \left( v_0 \gae{0} + v_1 \gae{1} + v_2 \gae{2} + v_3 \gae{3} + v_4 \gae{\infty} \right)^*.
                \end{equation}
                Note that if $X$ is a plane then $v_0=0$, otherwise $X$ is a sphere. Next, we want to project $P_a$ to $X$ by using the general subspace projection formula of CGA
                \begin{equation}\label{eq:projection_geometric_primitive}
                P_{pair} = \big((P_a \outer \gae{\infty}) \inner X\big) X^{-1}.
                \end{equation}
                Here we first construct the pointpair $P_a \outer \gae{\infty}$, where $\gae{\infty}$ corresponds to the point at infinity. $P_a \outer \gae{\infty}$ is also called a flat point. Note that the projection essentially amounts to first constructing the dual line $(P_a \outer \gae{\infty}) \inner X$ that passes through the point $P_a$ and is orthogonal to $X$, then intersecting this line with the primitive $X$.

                If $X$ is a sphere, then the intersection of the line and the sphere will result in two points on the sphere. If $X$ is a plane, it will result in another flat point, i.e. one point on the plane and one at infinity. In any case, we can retrieve the closer one to the agent position $P_a$ using the split operation
                \begin{equation}\label{eq:split_pointpair}
                    P_a' = \text{split}\left[P_{p} \right].
                \end{equation}
                Here, $P_a'$ is the projected agent position on the tangent space $X$. Next, we compute our agent's footprint (i.e.,\ instantaneous coverage) by projecting its surface to the point cloud. If the target surface was flat, all the points within the radius $r_a$ of our agent $P_a'$ would be covered by the footprint. However, in the general case, both the tool and the surface can be curved and deformable. For simplicity, we assume that the surface is rigid, and it deforms the tool with a constant bending radius. We use the radius of the local tangent sphere that we computed using CGA as an approximation for the bending radius. Accordingly, we can quantify the error of the local tangent space approximation for the $i$-th neighbor $P_i$ by the normalized residuals $e_i$ of the least squares computation~\eqref{eq:surface_fitting_minimization}. We encode this approximation error into the footprint by weighting the $i$-th neighbor by the Gaussian kernel $\varphi(r)$ using the normalized residuals $r_i = e_i/\max(\bm{e})$
                \begin{equation}{\label{eq:gaussian_kernel}}
                    \varphi(r_i)=\exp{\left(-\varepsilon^2r_i^2\right)},
                \end{equation}
                where the hyperparameter $\varepsilon>0$ controls the coverage falloff. Next, we substitute the Gaussian kernel weighted footprint into~\eqref{eq:coverage} to compute the coverage $\bm{c}_t$, which is then used to calculate the virtual source term $\bm{s}_t$ via~\eqref{eq:virtual_source}. As mentioned earlier, this virtual source term serves as the initial condition for the diffusion equation~\eqref{eq:heat} at each iteration of the tactile coverage loop, i.e.,\ $\bm{u}_0=\bm{s}_t$.

                

            \subsubsection{Gradient of the Diffused Potential Field}
            \label{sub:gradient_of_potential}
                We guide the coverage agent using the gradient of the diffused potential field as the acceleration command
                \begin{equation}{\label{eq:acceleration}}
                    \ddot{P}'_a = \nabla u_{P'_a,\diffusiontime},
                \end{equation}
                where $\nabla u_{P'_a,\diffusiontime}$ denotes the gradient of the diffused potential field at the projected agent position $P'_a$. However, computing the gradient on the point cloud is more involved than a regular grid or a mesh. Recall that in Section~\ref{sub:projection}, we already computed the projected agent position $P_a'$, the local neighborhood and the tangent space $X^*$. As the first step, we compute the tangent plane $E_{a,\diffusiontime}$ at $P_a'$, namely
                \begin{equation}\label{eq:tangent_plane}
                    E_{a,\diffusiontime} = \dual{L_{a,\perp}} \outer P_a' \outer \gae{\infty},
                \end{equation}

                using the line $L_{a,\perp}$, which is orthogonal to the surface and passes through $P_a'$. It is found by wedging the dual primitive $X$ with $P_a'$ to infinity with
                \begin{equation}\label{eq:orthogonal_line}
                    L_{a,\perp} = \dual{X} \outer P_a' \outer \gae{\infty}.
                \end{equation}

                Then, we project the points $P_i$ in the local neighborhood to the tangent plane $E_{a,\diffusiontime}$ using \eqref{eq:projection_geometric_primitive} and~\eqref{eq:split_pointpair}, by setting $E_{a,\diffusiontime}$ as the primitive $X$. Next, we use the values of the potential field at the neighbor locations as the height $h_i=u_{i,\diffusiontime}$ of a second surface from the tangent plane. Then, we fit a $3$-rd degree polynomial to this surface as shown by using the weighted least squares objective
                \begin{equation}
                    \hat{\bm{A}}=\arg\min_{\bm{A}} \operatorname{tr}\Big((\bm{Y}-\bm{X} \bm{A})^{\trsp} \bm{W}(\bm{Y}-\bm{X} \bm{A})\Big),   
                \end{equation}
                with the diagonal weight matrix $\bm{W}$
                \begin{equation}
                    \bm{W}=\diag\Big(\varphi(r_1), \varphi(r_1), \ldots \varphi(r_m)\Big),
                \end{equation}
                whose entries are given by the Gaussian kernel~\eqref{eq:gaussian_kernel}. One can refer to~\cite{nealenAsShortAsPossibleIntroductionLeast2004a} for the details. Lastly, we calculate the gradient at the projected agent's position using the analytical gradients of the polynomial. We depict the approach visually in Figure~\ref{fig:gradient}.

                \input{floats/gradient}

    \subsection{Robot Control}
    \label{sub:robot_control}
        There are several aspects that the control of the physical robot needs to achieve. The first is to track the virtual coverage agent on the target surface, while keeping the end-effector normal to the surface. The second is to exert a desired force on the surface. To do so, we design a task-space impedance controller while further exploiting geometric algebra for efficiency and compactness. The control law is of the following form
        \begin{equation}\label{eq:ga_task_space_impedance_control}
            \torquejoint = -\transpose{\gamatrix{J}} \inner \wrench,
        \end{equation}
        where $\gamatrix{J} \in \mathbb{B}^{1\times N}\subset \cga^{1\times N}$ is the Jacobian multivector matrix with elements corresponding to bivectors, $\wrench$ is the desired task-space wrench and $\torquejoint$ are the resulting joint torques. Before composing the final control law, we will explain its components individually.

        
        \subsubsection{Surface Orientation}
        \label{ssub:surface_orientation}
            From Equation~\eqref{eq:orthogonal_line}, we obtained a line $L_{a,\perp}$ that is orthogonal to the surface that we wish to track. In~\cite{lasenbyCalculatingRotorConformal2019}, it was shown how the motor between conformal objects can be obtained. We use this formulation to find the motor between the target orthogonal line and the line that corresponds to the $z$-axis of the end-effector of the robot in its current configuration, which is found as
            \begin{equation}\label{eq:end_effector_line}
                L_{ee} = M(\posjoint) (\gae{0} \outer \gae{3} \outer \gae{\infty}) \reverse{M}(\posjoint).
            \end{equation}
            Then, the motor $M_{L_{ee}L_{a,\perp}}$, which transforms $L_{ee}$ into $L_{a,\perp}$ can be found as
            \begin{equation}\label{eq:ga_line_motor}
                M_{L_{ee}L_{a,\perp}} = \frac{1}{C} \left( 1 + L_{a,\perp}L_{ee} \right),
            \end{equation}
            where $C$ is a normalization constant. Note that $C$ does not simply correspond to the norm of $1 + L_{a,\perp}L_{ee}$, but requires a more involved computation. We therefore omit its exact computation here for brevity and refer readers to~\cite{lasenbyCalculatingRotorConformal2019}. 

            We can now use the motor $M_{L_{ee}L_{a,\perp}}$ in order to find a control command for the robot via the logarithmic map of motors, i.e. 
            \begin{equation}\label{eq:line_control_command_twist}
                \twist_{L_{a,\perp}} = \log \left( M_{L_{ee}L_{a,\perp}} \right).
            \end{equation}
            Of course, if the lines are equal, $M_{L_{ee}L_{a,\perp}} = 1$ and consequently $\twist_{L_{a,\perp}} = 0$. Note that $\twist_{L_{a,\perp}}$ is still a command in task space (we will explain how to transform it to a joint torque command once we have derived all the necessary components).

            Another issue is that algebraically, $\twist_{L_{a,\perp}}$ corresponds to a twist, not a wrench. Hence, we need to transform it accordingly. From physics, we know that twists transform to wrenches via an inertial map, which we could use here as well. In the context of control, this inertia tensor is, however, a tuning parameter and does not actually correspond to a physical quantity. Thus, in order to simplify the final expression, we will use a scalar matrix valued inertia, instead of a geometric algebra inertia tensor and choose to transform the twist command to wrench command purely algebraically. As it has been shown before, this can be achieved by the conjugate pseudoscalar $I_c = I\gae{0}$~\cite{hestenesNewToolsComputational2010}. It follows that 
            \begin{equation}\label{eq:line_control_command}
                \wrench_{L_{a,\perp}} = \twist_{L_{a,\perp}} I_c,
            \end{equation}
            and $\wrench_{L_{a,\perp}}$ now algebraically corresponds to a wrench.

        
        \subsubsection{Target Surface Force}
        \label{ssub:target_surface_force}
            Since this article describes a method for tactile surface coverage, the goal of the robot control is to not simply stay in contact with the surface, but to actively exert a desired force on the surface. First of all, we denote the current measured wrench as $\wrench_m(t)$ and the desired wrench as $\wrench_d$. Both are bivectors as defined by Equation~\eqref{eq:wrench_space}. We use $\wrench_d$ w.r.t.\ end-effector in order to make it more intuitive to define. Hence, we need to transform $\wrench_m(t)$ to the same coordinate frame, i.e. 
            \begin{equation}\label{eq:measured_wrench_transformation}
                \wrench_m'(t) = \reverse{M}(\posjoint) \wrench_m(t) M(\posjoint).
            \end{equation}

            In order to achieve the desired interaction force, we simply apply a standard PID controller in wrench space, i.e.
            \begin{equation}\label{eq:wrench_pid}
                \wrench_C =  \bm{K}_{p,\wrench} \wrench_e + \bm{K}_{i,\wrench} \int_0^\trsp \wrench_e(\tau)\text{d}\tau + \bm{K}_{d,\wrench} \frac{\text{d} }{\text{d} t} \wrench_e(t),
            \end{equation}
            where the wrench error is 
            \begin{eqnarray}\label{eq:wrench_error_dynamics}
                \wrench_e(t) &=& \wrench_d - \wrench_m'(t),
            \end{eqnarray}
            where $\bm{K}_{p,\wrench}, \bm{K}_{i,\wrench}$ and $\bm{K}_{d,\wrench}$ are the corresponding gain matrices, and $\wrench_C$ is the resulting control wrench.

            Since the desired wrench is defined in end-effector coordinates, it usually amounts to a linear force in the $z$-direction of the end-effector frame, i.e. $\wrench_d = f_d \gae{03}$. Additionally, for an improved cleaning behavior one could also set a desired torque around that axis by adding $\tau_d \gae{12}$. The pattern of how to set this torque, however, would be subject to further investigation.

        
        \subsubsection{Task-Space Impedance Control}
        \label{ssub:task_space_impedance_control}
            Recalling the control law from Equation~\eqref{eq:ga_task_space_impedance_control}, we now collect the terms from the previous subsections into a unified task-space impedance control law. We start by looking in more detail at the Jacobian $\gamatrix{J}$. Previously, we mentioned that we are using the current end-effector motor as the reference, hence, we require the Jacobian to be computed w.r.t.\ that reference. This is therefore not the geometric Jacobian that was presented in Equation~\eqref{eq:ee_geometric_jacobian}, but a variation of it. The end-effector frame geometric Jacobian $\gamatrix{J}_G^{ee}$ can be found as 
            \begin{equation}\label{eq:ee_frame_geometric_jacobian}
                \gamatrix{J}_G^{ee} = \mat{
                    B_1^{ee}
                    & \ldots &
                    B_N^{ee}
                },
            \end{equation}
            where the bivector elements can be found as
            \begin{equation}\label{eq:frame_jacobian_bivector_element}
                B_i^{ee} = \reverse{M}_i^{ee}(\posjoint) B_i M_i^{ee}, 
            \end{equation}
            with 
            \begin{equation}\label{eq:jacobian_motor}
                M_i^{ee} = \prod_{j=N}^i M_i(q_i).
            \end{equation}
            Hence, the relationship between $\gamatrix{J}_G$ and $\gamatrix{J}_G^{ee}$ can be found as 
            \begin{equation}\label{eq:jacobians_relationship}
                \gamatrix{J}_G^{ee} = \reverse{M}(\posjoint) \gamatrix{J}_G M(\posjoint).
            \end{equation}
            
            The wrench in the control law is composed of the three wrenches that we defined in the previous subsections. As commonly done, we add a damping term that corresponds to the current end-effector twist and as before, we transform it to an algebraic wrench, i.e. 
            \begin{equation}\label{eq:damping_wrench}
                \wrench_{\twist} = \gamatrix{J}_G^{ee} \veljoint \gae{0\infty}.
            \end{equation}

            With this, we now have everything in place to compose our final control law as 
            \begin{equation}\label{eq:final_control_law}
                \torquejoint = -\gamatrix{J}_G^{ee,\trsp} \inner \left( \bm{K}_{L_{a,\perp}} \wrench_{L_{a,\perp}} - \bm{D}_{\twist} \wrench_\twist + \wrench_C \right), 
            \end{equation}
            where $\bm{K}_{L_{a,\perp}}$ is a stiffness and $\bm{D}_{\twist}$ a damping gain. 

%% file: floats/gradient.tex
\begin{figure}[ht]
        \centering
        \includegraphics[width=1\linewidth]{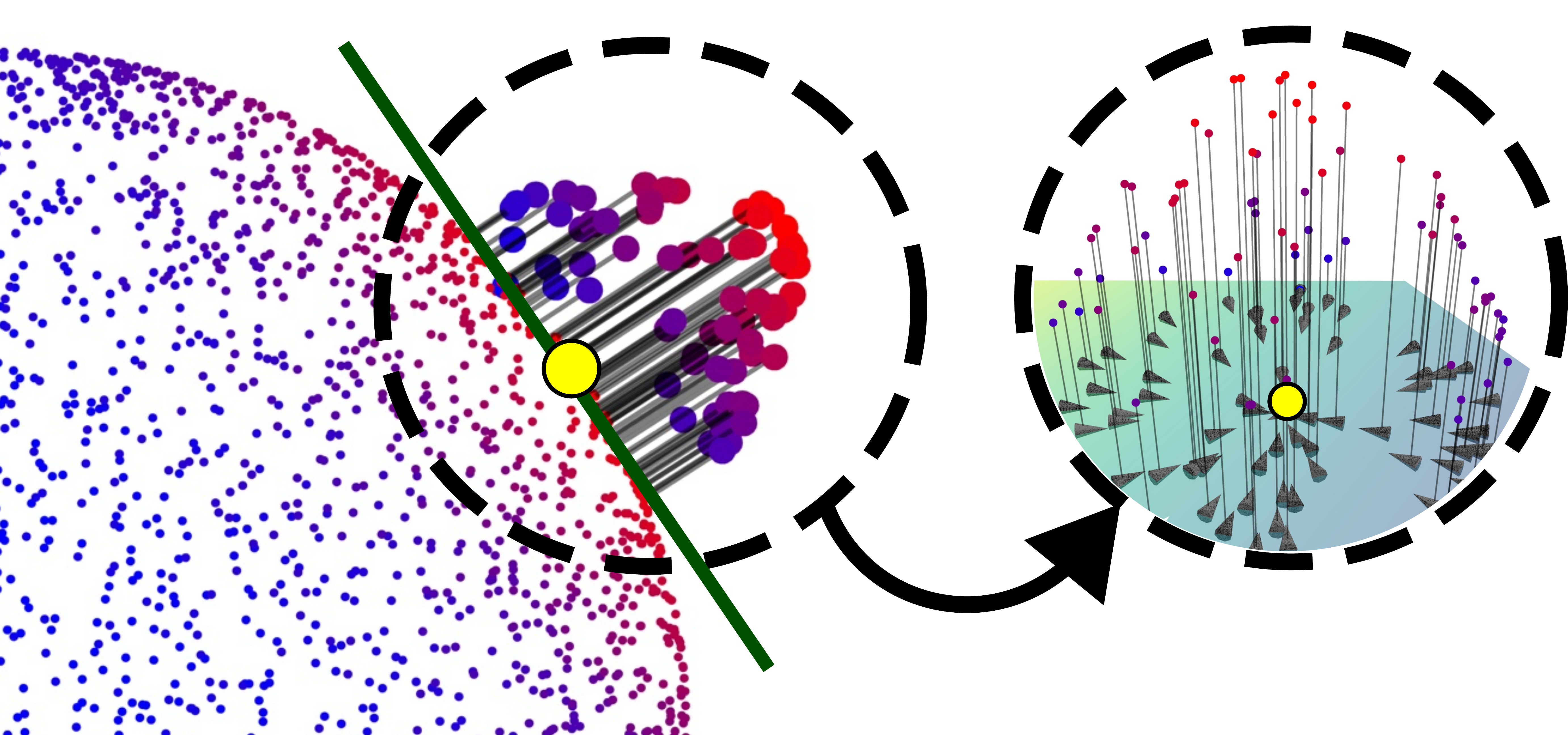}
        \caption{Blue-red points show the value of the potential field $u_\diffusiontime$ on the pointcloud $\mathcal{P}$ and the yellow point is the projected agent position $P'_a$. We also project the agent's neighbors $P_i$ to the tangent plane $E_{a,\diffusiontime}$, shown in green. Next, we use the height function $h_i=u_{i,\diffusiontime}$ which uses 
        the values of the potential field to lift the projected points in the normal direction of the tangent plane. We show the lifted points with large blue-red points. We fit a polynomial to this lifted surface and compute its analytical gradients at the neighbor locations $\nabla u_{i,\diffusiontime}$, as shown with arrows in the detail view.}
        \label{fig:gradient}
\end{figure}

%% file: sections/4Experiments.tex

\section{EXPERIMENTS}
\label{sec:experiments}

    Our experimental setup comprises a BotaSys SensOne 6-axis force torque (F/T) sensor attached to the wrist of a 7-axis Franka robot manipulator and a custom 3-D printed part attached to the F/T sensor. The custom part interfaces an Intel Realsense D415 depth camera and a sponge at its tip. We consider the sponge's center point to be the coverage agent's position $P_a$. Before the operation, we perform extrinsic calibration of the camera to combine the depth and RGB feeds from the camera and to obtain its transformation with respect to the robot joints. Additionally, we calibrate the F/T sensor to compensate for the weight of the 3-D printed part and the camera. We show the experimental setup on the left of Figure~\ref{fig:overview}.

    \subsection{Implementation Details}
    \label{sub:implementation_details}
        The pipeline of our tactile ergodic coverage method consists of three modules: (i) surface acquisition, (ii) surface coverage and (iii) robot control. Figure~\ref{fig:logic_flow} summarizes the information flow between the components.
        
        \input{floats/logic_flow}

        \subsubsection{Surface Acquisition} 
        \label{sub:surface_acquisition}
            The surface acquisition node is responsible for collecting the point cloud and performing preprocessing operations described in Section~\ref{sub:surface_preprocessing}. We use \emph{scipy}\footnote{\url{https://scipy.org}} for the nearest neighbor queries and for solving the eigenproblem in \eqref{eq:eigenproblem}. The matrices $\bm{C}$ and $\bm{M}$ composing the discrete Laplacian in \eqref{eq:discrete_Laplacian} are computed with the \emph{robust\_laplacian} package\footnote{\url{https://github.com/nmwsharp/robust-laplacians-py}}~\cite{sharpLaplacianNonmanifoldTriangle2020}.

        \subsubsection{Surface Coverage} 
        \label{sub:surface_coverage}
            The surface coverage node performs the computations based on the procedure given in Section~\ref{sub:tactile_coverage}. It uses the information provided by the surface acquisition node and produces the target \emph{line} for the robot control node.

        \subsubsection{Robot control} 
        \label{sub:robot_control_experiment}
            On a high level, the robot control can be seen as a state machine with three discrete states. The first two states are essentially two pre-recorded joint positions in which the robot is waiting for other parts of the pipeline to be completed. One of these positions corresponds to the picture-taking position, i.e.,\ a joint position where the camera has the full object in its frame and the point cloud can be obtained. The robot is waiting in this position until the point cloud has been obtained, afterwards it changes its position to hover shortly over the object. In this second position, it is waiting for the computation of the Laplacian eigenfunctions to be completed, such that the coverage can start. The switching between those two positions is achieved using a simple joint impedance controller. 

            The third and most important state is when robot is actually controlled to be in contact with the surface and to follow the target corresponding to the coverage agent. This behaviour is achieved using the controller that we described in Section~\ref{sub:robot_control}. The relevant parameters, that were chosen empirically for the real-world experiments, are the stiffness and damping of the line tracking controller, i.e. $K_{L_{a,\perp}} = \text{diag}(30, 30, 30, 750, 750, 300)$ and $D_\twist=\text{diag}(10, 10, 10, 150, 150, 50)$, as well as the gains of the wrench PID controller, i.e. $K_{p,\wrench}=0.5$, $K_{i,\wrench}=5$ and $K_{d,\wrench}=0.5$. The controller has been implemented using our open-source geometric algebra for robotics library \textit{gafro}\footnote{\url{https://gitlab.com/gafro}} that we first presented in~\cite{lowGeometricAlgebraOptimal2023}. Note that in some cases, matrix-vector products of geometric algebra quantities have been used for the implementation, where the mathematical structure of the geometric product actually simplifies to this, which can be exploited for more efficient computation.

    \subsection{Simulated Experiments}
    \label{subsec:simulated_experiments}
        \subsubsection{Computation Performance}
        \label{subsec:computation_performance}
            In order to assess the computational performance, we investigated the two main operations of our method: (i) preprocessing by solving either the eigenproblem~\eqref{eq:eigenproblem} or matrix inversion in~\eqref{eq:implicit_time_stepping} (ii) integrating the diffusion at runtime using either the spectral~\eqref{eq:spectral_time_stepping} or implicit~\eqref{eq:implicit_time_stepping} formulations. In this experiment, we used the Stanford Bunny as the reference point cloud and performed voxel filtering to set the point cloud resolution. We present the results for the preprocessing in Figure~\ref{fig:complexity_preprocessing} and for the runtime in Figure~\ref{fig:complexity_integration}.

            \input{floats/complexity_preprocessing}
            \input{floats/complexity_integration}

        \subsubsection{Coverage Performance}
        \label{subsec:coverage_performance}
            We tested the coverage performance in a series of kinematic simulations. As the coverage metric, we used the normalized ergodicity over the target distribution, which compares the time-averaged statistics of agent trajectories to the target distribution 
            \begin{equation}
                \bm{\varepsilon}_t = \frac{\|\max\left(\bm{p}-\bm{c}_t,0\right)\|_2}{\sum_{i=1}^{\npoints}p_i}.
                \label{eq:ergodicity}
            \end{equation}
            

            We ran the experiments for three different objects: a partial point cloud of the Stanford Bunny and two point clouds of a cup and a plate and their target distributions that we collected using the RGB-D camera. For the Stanford bunny, we projected an `X' shape as the target distribution. For each object, we sampled ten different initial positions for the coverage agent and kinematically simulated the coverage using different numbers of eigencomponents $\neigen = 25,50,100,200$ and diffusion timestep scalar $\alpha=1,5,10,50,100$. Since the plate is larger compared to the Bunny and the cup, we used a larger agent radius $r_a=15\ [mm]$ for the plate and a smaller value $r_a=7.5\ [mm]$ for the cup and the Bunny. The other parameters that we kept constant in all of the experiments are $\ddot{x}_{\text{max}}=3\ [mm/s^2]$, $\dot{x}_{\text{max}}=3\ [mm/s]$. We selected six representative experiment runs to show the coverage performance qualitatively, and present them in Figure~\ref{fig:simulation}.

            \input{floats/simulation}

            We show the quantitative results with respect to $\neigen$ and $\alpha$ in Figures~\ref{fig:boxplot_nb_eigen_filtered} and~\ref{fig:boxplot_alpha_filtered}, respectively. Note that, in order to better show performance trend in these plots, we have excluded parameter combinations leading to failure cases. We will discuss those in Section~\ref{sec:discussion}.
                
            \input{floats/boxplot_nb_eigen_filtered}
            \input{floats/boxplot_alpha_filtered}

            As the last experiment, we chose the best-performing pair $(\neigen,\alpha)$ and show the time evolution of the coverage performance for different objects in Figure~\ref{fig:shaded_plot_objects}.

            \input{floats/shaded_plot_objects}
    
    \subsection{Real-world Experiment}
    \label{subsec:real_world_experiment}
        In the real-world experiments, we tested the whole pipeline presented in Section~\ref{sub:implementation_details}. We used three different kitchen utensils (plate, bowl, and cup) with different target distributions (shapes, RLI, X). For these experiments, we fixed the objects to the table so that they could not move when the robot was in contact. At the beginning of the experiments, we moved the robot to a predefined joint configuration that fully captured the target distribution. Since we collected the point cloud data from a single image frame, our method only had access to a partial and noisy point cloud. We summarize the results of the real-world experiments in Figure~\ref{fig:real_world_experiment} and share all the recorded experiment data and the videos on the accompanying website.
        \input{floats/real_world_experiment}
    \subsection{Comparisons}
    \label{subsec:comparisons}
        We present the first tactile ergodic coverage method in the literature that works on curved surfaces. Therefore, there are no methods that we can directly compare to quantitatively. For this reason, we selected three related state-of-the-art methods and compared them to our method qualitatively. As the first method, we selected the finite element based HEDAC planner~\cite{ivicMultiUAVTrajectoryPlanning2023}, since it is the only other ergodic control approach working on curved surfaces. For the tactile interaction aspect, we selected two methods, the unified force-impedance control~\cite{karacanTactileExplorationUsing2023} and the sampling-based informative path planner~\cite{kabirAutomatedPlanningRobotic2017}. We specified six criteria for comparison and summarized the results in Table~\ref{tab:comparison}.

\input{floats/comparison_table} 

%% file: floats/logic_flow.tex
\begin{figure*}[ht]
        \centering
        \includegraphics[width=0.8\linewidth]{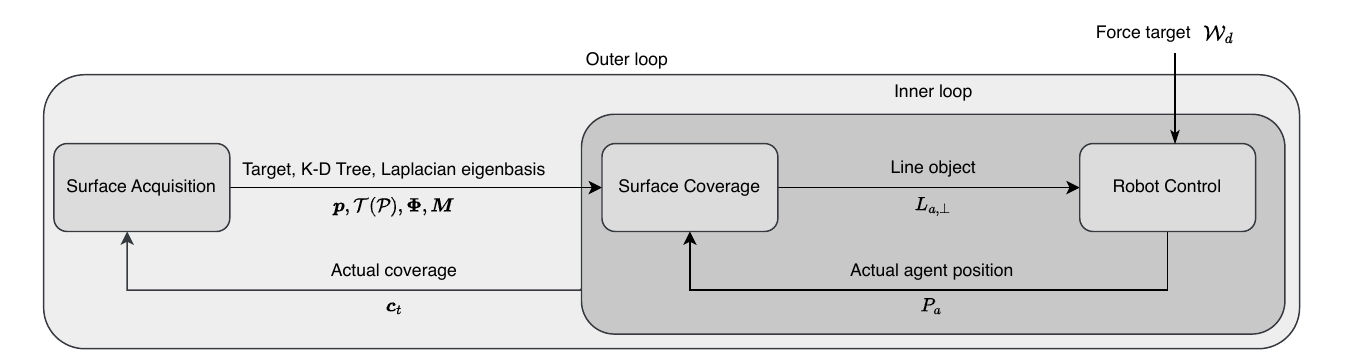}
        \caption{Information flow between the three components. The pipeline is composed of an outer loop responsible for controlling the coverage progress with the feedback from the camera, whereas the inner loop compensates for the mismatch due to the robot dynamics.}
        \label{fig:logic_flow}
\end{figure*}

%% file: floats/complexity_preprocessing.tex
\begin{figure}[ht]
        
        \centering
        \includegraphics[width=1\linewidth]{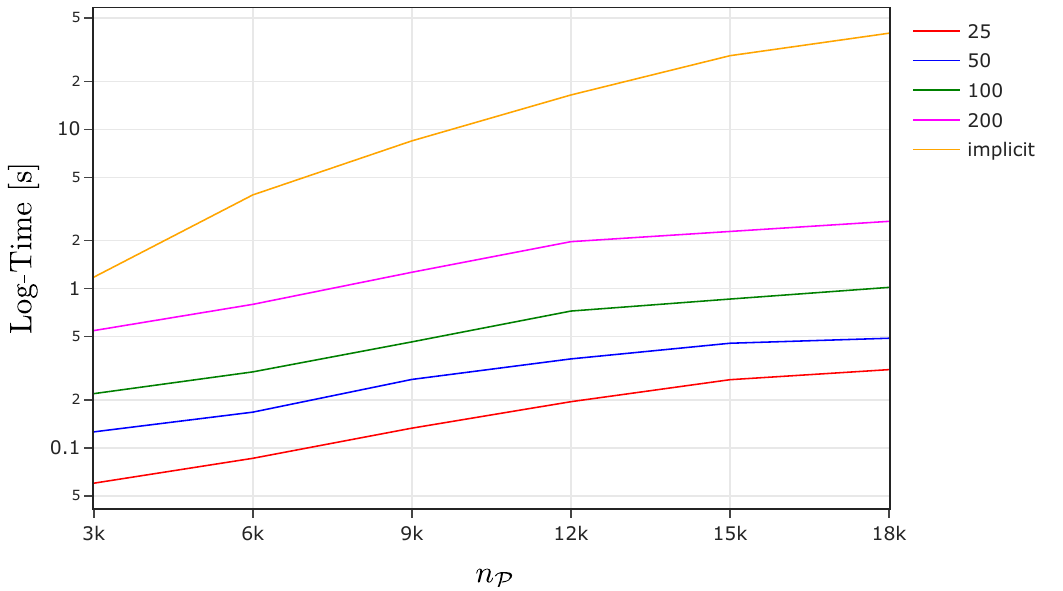}
        \caption{Computational complexity of the preprocessing step for different $\npoints$ and $\neigen$. Legend shows $\neigen$ values. The time axis is logarithmic and the legend shows $\neigen$ values.}
        \label{fig:complexity_preprocessing}
\end{figure}

%% file: floats/complexity_integration.tex
\begin{figure}[ht]
        \centering
        \includegraphics[width=1\linewidth]{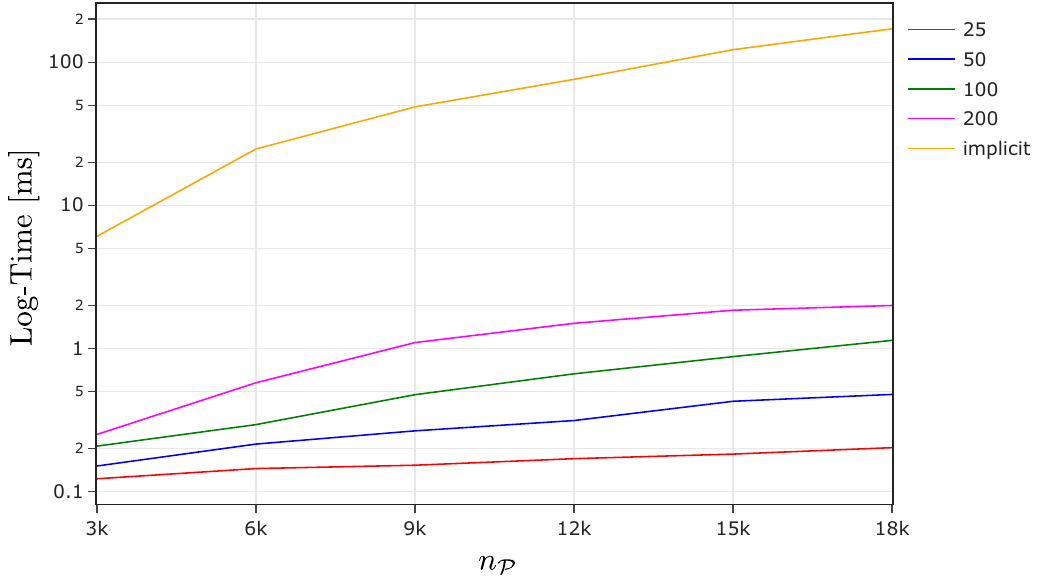}
        \caption{Computational complexity of integrating the diffusion equation at runtime for different $\npoints$ and $\neigen$. The time axis is logarithmic and the legend shows $\neigen$ values.}
        \label{fig:complexity_integration}
\end{figure}

%% file: floats/simulation.tex
\begin{figure*}[ht]
        \centering
        \includegraphics[width=1\linewidth]{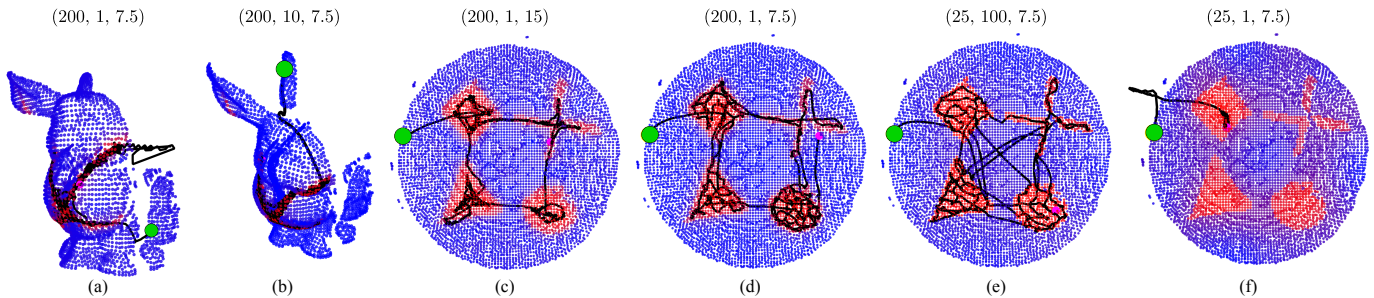}
        \caption{Qualitative results of the coverage experiments showcasing the effect of different parameters. The red points designate the spatial target distribution $p_i>0$. The agent starts at the green point, the trajectory is shown in black, and the final position after 1000 timesteps is shown with the purple point. The tuples given on top of the figures show the parameters $n_{\mathcal{K}}$, $\alpha$, and $r_a$ of the experiments. We provide the interactive point clouds and the experiment data on our website.}
        \label{fig:simulation}
\end{figure*}

%% file: floats/boxplot_nb_eigen_filtered.tex
\begin{figure}[ht]
        \centering
        \includegraphics[width=1\linewidth]{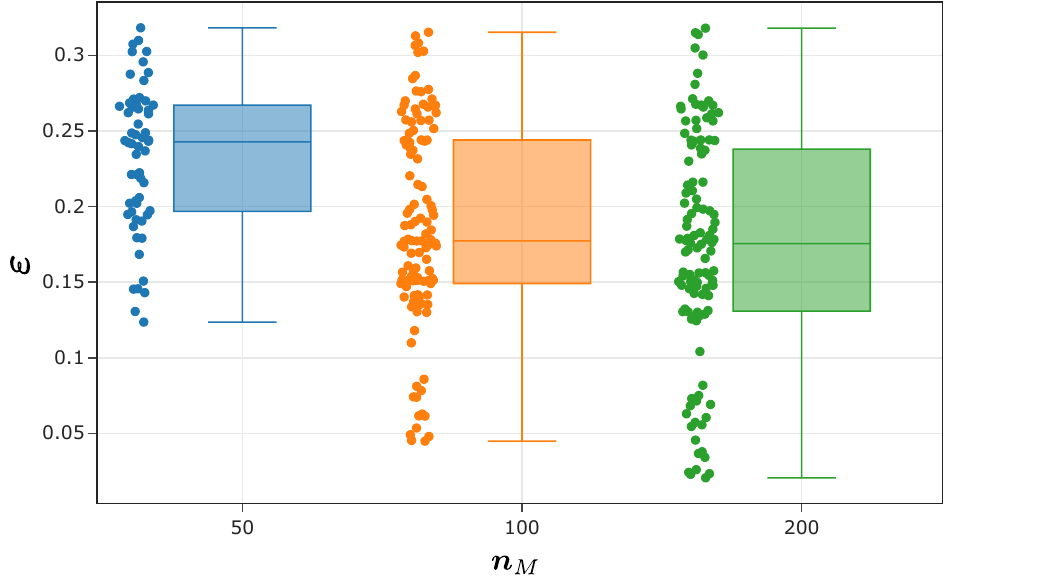}
        \caption{Coverage performance measured by the ergodic metric $\bm{\varepsilon}_t$ \eqref{eq:ergodicity} with respect to $\neigen$ used in the spectral formulation \eqref{eq:spectral_diffusion}.}
        \label{fig:boxplot_nb_eigen_filtered}
\end{figure}

%% file: floats/boxplot_alpha_filtered.tex
\begin{figure}[ht]
        \centering
        \includegraphics[width=1\linewidth]{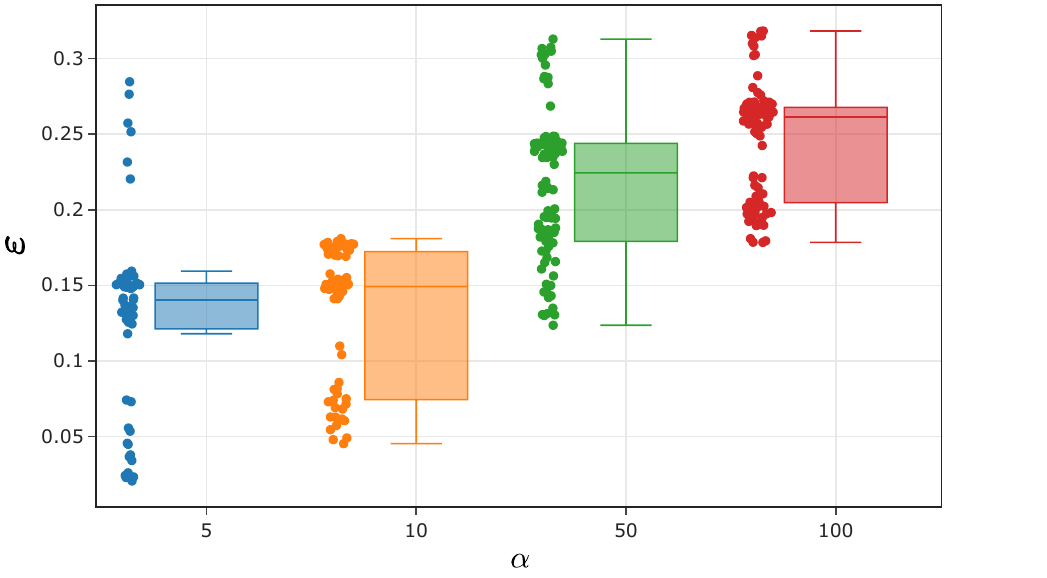}
        \caption{Coverage performance measured by the normalized ergodic metric $\bm{\varepsilon}_t$ \eqref{eq:ergodicity} with respect to the parameter $\alpha$.}
        \label{fig:boxplot_alpha_filtered}
\end{figure}

%% file: floats/shaded_plot_objects.tex
\begin{figure}[ht]
        \centering
        \includegraphics[width=1.0\linewidth]{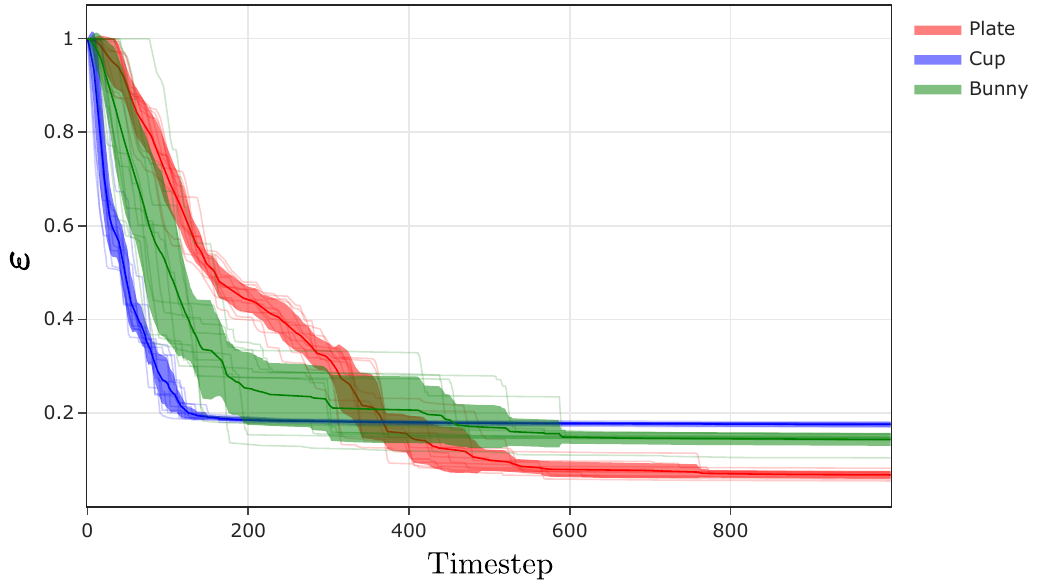}
        \caption{Time evolution of the ergodic metric \eqref{eq:ergodicity} for three different objects with $\neigen=200$ and $\alpha=10$. The semi-transparent lines show ten different experiment runs, the center line shows the mean, and the shaded regions correspond to the standard deviation.}
        \label{fig:shaded_plot_objects}
\end{figure}

%% file: floats/real_world_experiment.tex
\begin{figure}[ht]
        \centering
        \includegraphics[width=1.0\linewidth]{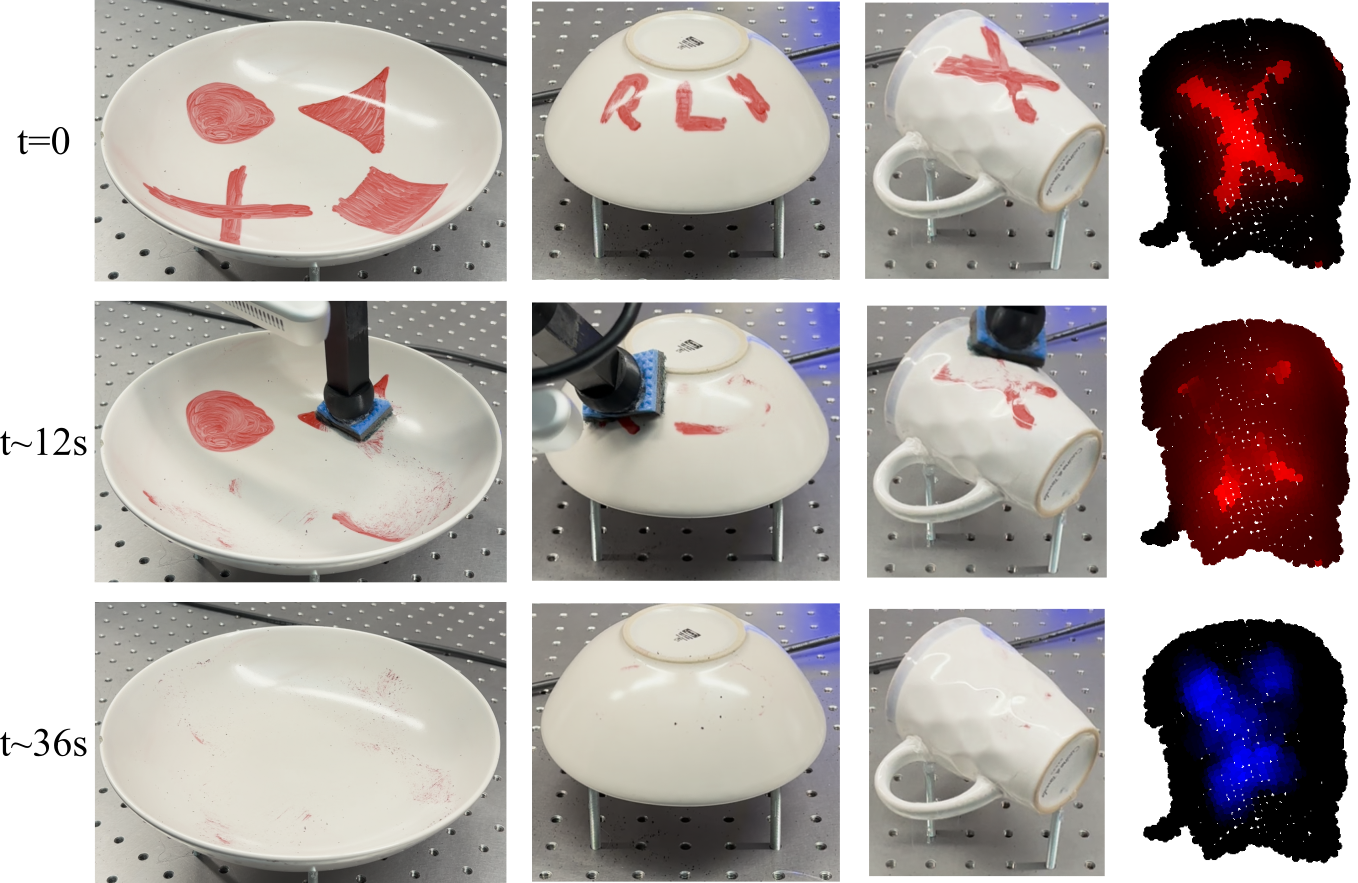}
        \caption{Real-world experiment of the robot cleaning a plate, a bowl, and a cup. For the first three columns we give snapshots from the initial, intermediate, and final states from top to the bottom. In the last column, we show the target distribution $\bm{p}$, the simulated potential field $\bm{u}_t$ and the coverage $\bm{c}_t$ from top to the bottom. 
        }
        \label{fig:real_world_experiment}
\end{figure}

%% file: floats/comparison_table.tex
\begin{table*}[ht]
    \centering
    \begin{minipage}{0.85\linewidth} 
        \caption{Comparison of the proposed method with state-of-the-art methods.}
        \label{tab:comparison}
        \begin{tabular}{lllllll}
            \toprule
                Method
                & Domain      
                & Approach 
                & Online 
                & Purpose 
                & Multiscale
                & Multisetup \footnote{Multisetup used by~\cite{kabirAutomatedPlanningRobotic2017} refers to planning the configuration of the target object to reach otherwise unreachable regions.} 
                \\
            \midrule
                Finite element-based HEDAC~\cite{ivicMultiUAVTrajectoryPlanning2023}      
                & Mesh        
                & Planning         
                & No     
                & Visual Inspection    
                & Yes        
                & No         
                \\ 
                Sampling-based Planner~\cite{kabirAutomatedPlanningRobotic2017}  
                & Mesh        
                & Planning         
                & Yes    
                & Tactile Coverage     
                & No         
                & Yes        
                \\
                Unified Force-Impedance Control~\cite{karacanTactileExplorationUsing2023}  
                & None        
                & Control
                & Yes
                & Surface Exploration
                & No         
                & No        
                \\
                Tactile Ergodic Control (Ours)           
                & Point Cloud \footnote{Since point clouds are the most general representation, our methoud can seamlessly be used on grids/meshes with only minor changes to the computation of the discrete Laplacian.}
                & Control          
                & Yes    
                & Tactile Coverage   
                & Yes        
                & No         
                \\ 
            \bottomrule
        \end{tabular}
        \vspace{-1em}
    \end{minipage}
        
\end{table*}


%% file: sections/5Conclusion.tex

\section{DISCUSSION}
\label{sec:discussion}
    
    \subsection{Computational Performance}
    \label{sub:computational_performance}
        We investigated the computational performance of our method for the preprocessing and for the runtime.

        
        The preprocessing step is only required, when the robot sees an object for the first time or when the object undergoes a non-isometric transformation. First thing to note from Figure~\ref{fig:complexity_preprocessing} is that computing the eigenbasis is significantly faster than inverting the large sparse matrix. Secondly, the advantage of the spectral approach becomes more significant as the number of points increases. This is because the computational complexity of the spectral approach is linear $\mathcal{O}(\npoints\neigen)$ with the number of points, whereas the matrix inversion of the implicit solution has quadratic complexity $\mathcal{O}(\npoints^2)$.
        
        If we compare our method with the state-of-the-art in ergodic coverage on curved surfaces~\cite{ivicMultiUAVTrajectoryPlanning2023}, our preprocessing step is significantly faster. They reported a computation time of \SI{19.7}{\second} for a mesh with 2315 points using a finite-element-based method. In contrast, our method takes \SI{278}{\milli\second} for a point cloud with $\approx 3000$ points with $\neigen=100$. Therefore, in comparison, our method promises an increase in computation speed of more than $90$ times. Note that, as the number of points increases, our gains in computation time become even more significant due to the difference in the computational complexity of the spectral and implicit formulations as mentioned above. 
        
        As Figure~\ref{fig:complexity_integration} shows the spectral approach also results in a significant performance increase at runtime. The implicit solution is also efficient in runtime, since it reduces to matrix-vector multiplication after inverting the sparse matrix at the preprocessing step. Nevertheless, the spectral formulation is still significantly faster than the implicit formulation, especially for large point clouds. Obviously, an unnecessarily large eigenbasis for small point clouds, i.e. $\neigen \to \npoints$, would cause the spectral approach to be slower than the implicit one.

    
    \subsection{Coverage Performance}
    \label{sub:coverage_performance}
        A close investigation of the failure scenarios in Figure~\ref{fig:simulation} revealed that they stem from the bad coupling of the parameters and from an initialization of the agent far away from the source. If the agent is not far away from the source, setting low values for $\alpha$ might actually lead to desirable properties such as prioritizing local coverage which would in turn minimize the distance traveled during coverage. Hence, for getting the best behavior, $\alpha$ can be set adaptively or sequentially. For instance, it is better to use high $\alpha$ values at the start for robustness to bad initializations and to decrease it as the coverage advances to prioritize local coverage and to increase the performance.

        We measured the effect of our method parameters on the coverage performance in Figures~\ref{fig:boxplot_nb_eigen_filtered} and~\ref{fig:boxplot_alpha_filtered}. Interestingly, the parameters influencing the agent's speed, i.e. $\dot{\bm{x}}_{max}$, $\neigen$ and $\alpha$, have a coupled effect on the coverage performance in some of the scenarios. The first thing to note here is that the value of the $\alpha$ is lower-bounded by the speed of the coverage agent $\dot{\bm{x}}_{max}$. Otherwise, the method cannot guide the agent since it moves faster than the diffusion. For instance, we observe from Figure~\ref{fig:simulation} a) and f) that with a diffusion coefficient $\alpha=1$, the source information does not propagate fast enough to the agent if it is too far from the source. Even for a small eigenbasis $\neigen\leq50$ and moderate diffusion coefficient values $1 < \alpha \leq 10$, it still results in a low coverage performance $\bm{\varepsilon}_t>0.5$. On the contrary, if the eigenbasis is chosen to be sufficiently large $\neigen\geq100$, we have more freedom in choosing $\alpha$. 

        With this in mind, we removed the infeasible parameter combinations $(\neigen=50, \alpha=\{5, 10\})$ from the experiment results in Figures~\ref{fig:boxplot_nb_eigen_filtered} and~\ref{fig:boxplot_alpha_filtered} to better observe the performance trend for $\neigen$ and $\alpha$. It is easy to see that increasing $\neigen$ results in increased performance and higher freedom in choosing $\alpha$. However, this benefit becomes marginal after $\neigen\geq100$. Therefore, choosing $\neigen=100$ becomes a good trade-off between coverage performance and computational complexity. This observation is in line with the value of $\neigen=128$ reported in~\cite{sharpDiffusionNetDiscretizationAgnostic2022}. 

        In Figure~\ref{fig:boxplot_alpha_filtered}, however, we observed minor differences in performance for different $\alpha$. Considering the spread and the mean, choosing $\alpha=10$ would be a good fit for most scenarios. Nevertheless, we must admit that the ergodic metric falls short in distinguishing the most significant differences between $\alpha$ values. Hence, the qualitative performance shown in Figure~\ref{fig:simulation} becomes much more explanatory. The first thing to note here is that the lower values of $\alpha$ result in more local coverage, whereas higher values lead to prioritizing global coverage. Accordingly, the tuning of this parameter depends on the task itself. For example, suppose the goal is to collect measurements from different modes of a target distribution as quickly as possible, in which case we would recommend using $\alpha > 50$. On the other hand, if the surface motion is costly, because for example, the surface is prone to damage, moving less frequently between the modes can be achieved by setting $5 < \alpha < 50$. 

        In scenarios where the physical interactions are complex, stopping the coverage prematurely and observing the actual coverage might be preferable instead of continuing the coverage. To decide when to actually pause and measure the current coverage, we investigated the time evolution of the coverage performance in Figure~\ref{fig:shaded_plot_objects}. For the cup and the bunny, we see that the coverage reaches a steady state around the $200$-th timestep, while for the plate, this occurs around the $500$-th timestep. Still, we can identify the steepest increase in the coverage occurring until the $150$-th timestep. Accordingly, we recommend the strategy to pause the coverage at roughly $200$ timesteps, measure the actual coverage, and continue the coverage. This would potentially help in the cases where we have unconnected regions (various modes), because discontinuous jumps between the disjoint regions might be quicker and easier than following the surface. All that said, these claims require further testing and experimentation, which are left to be investigated in future work.

    
    \subsection{Force Control}
    \label{sub:force_control}
        We demonstrated that the proposed method can perform closed-loop tactile ergodic control in the real world with unknown objects and target distributions, as depicted in Figure~\ref{fig:real_world_experiment}. The primary challenge, however, is to be keeping in contact with the surface without applying excessive force. This is mainly due to the insufficient depth accuracy of the camera, and uncertain dimensions of the mechanical system. A suboptimal solution is to use a compliant controller and adjust the penetration depth of the impedance target. A too compliant controller would, however, reduce the tracking precision and the uncertainty in the penetration depth could lead to unnecessarily high contact forces that might damage the object. More importantly, high contact forces result in high friction that further reduces the reference tracking performance.

        Our solution to this problem was to introduce tactile feedback from the wrist-mounted force and torque sensor and closed-loop tracking of a reference contact force. In general, the commands generated by the force controllers conflict with the position controllers and result in competing objectives. We overcome this problem by posing the objective as line tracking instead of position tracking. This forces the agent to be on the line but free to move along the line. Accordingly, the force and the line controller can simultaneously be active without conflicting objectives or rigorous parameter tuning.

    \subsection{Comparisons}
    \label{sub:comparisons_discussion}

    We compared our method with state-of-the-art approaches in Table~\ref{tab:comparison}. Since the methods are not comparable in all aspects, we discuss the advantages and disadvantages of our method in three parts: (i) ergodic coverage; (ii) tactile interactions; and (iii) tactile coverage. 
        
        
        \subsubsection{Ergodic Coverage}
        \label{ssub:ergodic_coverage}
            In the literature, the only other ergodic coverage method on curved surfaces is the finite element-based HEDAC~\cite{ivicMultiUAVTrajectoryPlanning2023}. This work presents an offline planning method on meshes for visual inspection using multiple aerial vehicles. Accordingly, our method extends the state of the art in ergodic coverage on curved surfaces by being the first formulation (i) working on point clouds, (ii) providing closed-loop coverage using vision, and (iii) performing tactile coverage. Furthermore, as we showed in the experiments in Section \ref{subsec:computation_performance}, our approach vastly outscales the finite element-based HEDAC in terms of computation time for the preprocessing step. It is also important to note that, due to the generality of the underlying ergodic control formulation that we are using, our method could be applied to their use-case as well. 

        
        \subsubsection{Tactile Interactions}
        \label{ssub:tactile_interactions}
            To ensure contact during tactile exploration, the usage of a unified force-impedance control scheme was proposed \cite{karacanTactileExplorationUsing2023}. The general idea is similar to ours, in the sense that the controller is required to track a given reference while exerting a force on the surface. The main difference stems from the formulation of the reference for the impedance behavior. While their method tracks a full Cartesian pose, our impedance controller tracks a line. The main difference here is that our method  imposes fewer constraints on the reference tracking, which leaves more degrees of freedom for secondary tasks, such as tracking the force objective. Hence, we require no additional tuning to integrate these objectives, whereas their method uses a passivity-based design to ensure the stability of the combined controller. 

        \subsubsection{Tactile Coverage}
        \label{ssub:comparisons_tactile_coverage}
            Concerning the problem of using a manipulator for tactile coverage on curved surfaces, we compare our method to the online sampling-based planner presented in~\cite{kabirAutomatedPlanningRobotic2017}. Unlike the more general point cloud representation that we are using, this method operates on meshes. However, it includes the planning of the configuration of the target object. This is currently a limitation of our approach, since we assume the object to be fixed and consider only a single viewpoint. However, their configuration planner is independent of the coverage at a given configuration, and it could be easily combined with our method.
            In contrast to our myopic feedback controller, their approach relies on trajectory planning, which requires a predefined planning horizon specified by the number of passes parameter to cover discrete patches. For tactile coverage tasks, this can be extremely challenging to estimate beforehand. Our method does not suffer from this limitation, since ergodicity guarantees revisiting continuous areas according to the target distribution over an infinite time horizon.
            In addition, their approach is based on generating splines that connect the waypoints. This has two issues: if the points are not densely sampled, there is no guarantee that the resulting spline would be on the surface; and conversely, if the points are densely sampled, then the spline would be very complex and not smooth. Accordingly, this approach would not scale to complex surfaces and target distributions. Our approach, on the other hand, uses a feedback controller to stay in contact with the surface, where the local references are coming from the surface-constrained ergodic controller. Hence, our approach is mainly limited by the robot's geometry with respect to the complexity of the object, which could also be mitigated by changing its configuration online.
     




\section{CONCLUSION AND FUTURE WORK}
\label{sec:conclusion}
    We presented the first closed-loop ergodic coverage method on point clouds for tactile coverage tasks on curved surfaces. Tactile coverage tasks are challenging to model due to complex physical interactions. We used vision to jointly capture the surface geometry and the target distribution as a point cloud and directly used this representation as input. Then, we propagated the information regarding the coverage target to our robot using a diffusion process on the point cloud. Here, we used ergodicity to relate the spatial distribution to the number of visits required for coverage in an infinite-horizon formulation. We leveraged a spectral formulation to trade-off the accuracy of the diffusion computation with its computational complexity. To find a favorable compromise between the two, we tested the dependency of the coverage performance to the hyperparameters in kinematic simulation experiments. Next, we demonstrated the method in a real-world setting by cleaning previously unknown curved surfaces with arbitrary human-drawn distributions. We observed that our method can  adapt and generalize to different object shapes and distributions on the fly. 


    
    Additionally, in some tactile coverage scenarios, it is not straightforward to measure the actual coverage using an RGB-D camera such as surface inspection, sanding, or mechanical palpation. Still, we can use cleaning as a proxy task such that a human expert can mark the regions that need to be inspected with an easy-to-remove marker. Then, the robot's progress would be detectable by a camera. Accordingly, our method provides an interesting human-robot interaction modality using annotations and markings of an expert for tactile robotics tasks.

    As discussed in Section~\ref{ssub:comparisons_tactile_coverage}, a practical limitation of our setup is fixing the object pose during the operation. Therefore, we plan to extend our method to scenarios where the object is grasped by a second manipulator and can be reconfigured for covering regions that otherwise would be unreachable due to either collisions or joint limits. Although this problem is easy to address by sampling discrete configurations, as previously done in~\cite{kabirAutomatedPlanningRobotic2017}, our goal is to extend our method to handle this problem in a continuous manner using a control approach.

    Another promising extension of our method is automating the collection of visuotactile datasets. In this setting, one can combine our method with a vision-based active learning module such as~\cite{ketchumActiveExplorationRealTime2024a}, which estimate high tactile-information regions on the surface. Then, our controller could be used to collect data from these regions with a multi-modal tactile sensor.